\documentclass{article}

\usepackage{microtype}
\usepackage{graphicx}
\usepackage{subfigure}
\usepackage{booktabs} 
\usepackage{paralist}
\usepackage{amssymb}

\usepackage{hyperref}
\usepackage{amsfonts} 
\usepackage{amsmath} 
\usepackage{bbm}
\usepackage{array,multirow,graphicx}
\usepackage{bm}
\usepackage{etoolbox,xspace}
\usepackage{footmisc}
\usepackage{algorithm}
\usepackage{algorithmic}  

\usepackage{hyperref}
\usepackage{xcolor}
\usepackage{booktabs}

\usepackage[accepted]{icml2020}

\icmltitlerunning{A Large-scale Study on Unsupervised Outlier Model Selection: Do Internal Strategies Suffice? }

\begin{document}

\twocolumn[
\icmltitle{A Large-scale Study on Unsupervised Outlier Model Selection: Do Internal Strategies Suffice? }



\icmlsetsymbol{equal}{*}

\begin{icmlauthorlist}
\icmlauthor{Martin Q. Ma}{equal,to}
\icmlauthor{Yue Zhao}{equal,to}
\icmlauthor{Xiaorong Zhang}{to}
\icmlauthor{Leman Akoglu}{to}
\end{icmlauthorlist}

\icmlaffiliation{to}{Carnegie Mellon University}

\icmlcorrespondingauthor{Martin Q. Ma}{qianlim@cmu.edu}
\icmlcorrespondingauthor{Yue Zhao}{zhaoy@cmu.edu}

\icmlkeywords{Machine Learning, ICML}

\vskip 0.3in
]

\newsavebox\CBox
\def\textBF#1{\sbox\CBox{#1}\resizebox{\wd\CBox}{\ht\CBox}{\textbf{#1}}}

\renewcommand{\algorithmicrequire}{\textbf{Input:}}
\renewcommand{\algorithmicensure}{\textbf{Output:}}
\renewcommand{\algorithmiccomment}[1]{\hfill$\blacktriangleright$ #1}

\newcommand{\cdash}{\multicolumn{1}{c}{--} }

\newcommand{\super}{$^\blacktriangle$}
\newcommand{\sub}{$^\triangledown$}

\newcommand{\qap}{$\bm{q}_{\textbf{AP}}$}
\newcommand{\qroc}{$\bm{q}_{\textbf{ROC}}$}
\newcommand{\qprn}{$\bm{q}_{\textbf{Prec}}$}

\newcommand{\method}{{\sc XX}}

\newcommand*{\bord}{\multicolumn{1}{c|}{}}

\newcommand{\cbit}{\begin{compactitem}}
	\newcommand{\ceit}{\end{compactitem}}
\newcommand{\cben}{\begin{compactenum}}
	\newcommand{\ceen}{\end{compactenum}}

\definecolor{OliveGreen}{rgb}{0,0.6,0}
\newcommand{\tick}{\textcolor{OliveGreen}{\ding{52}}}

\newcommand{\bit}{\begin{itemize}}
	\newcommand{\eit}{\end{itemize}}
\newcommand{\ben}{\begin{enumerate}}
	\newcommand{\een}{\end{enumerate}}
\newcommand{\beq}{\begin{equation}}
\newcommand{\eeq}{\end{equation}}

\newcommand{\ap}{AP\xspace}
\newcommand{\roc}{ROC\xspace}
\newcommand{\pre}{Prec@$k$\xspace}

\newcommand{\std}{{\sc std}\xspace}
\newcommand{\rs}{{\sc rs}\xspace}
\newcommand{\h}{{\sc h}\xspace}
\newcommand{\ch}{{\sc ch}\xspace}
\newcommand{\iind}{{\sc i}\xspace}
\newcommand{\dunn}{{\sc d}\xspace}
\newcommand{\s}{{\sc s}\xspace}
\newcommand{\db}{{\sc db}\xspace}
\newcommand{\xb}{{\sc xb}\xspace}
\newcommand{\sd}{{\sc sd}\xspace}

\newcommand{\emmes}{{\sc EM}\xspace}
\newcommand{\mv}{{\sc MV}\xspace}

\newcommand{\ireos}{{\sc IREOS}\xspace}
\newcommand{\udr}{{\sc UDR}\xspace}
\newcommand{\mc}{{\sc MC}\xspace}
\newcommand{\mcs}{{\sc MC$_S$}\xspace}

\newcommand{\hits}{{\sc HITS}\xspace}
\newcommand{\ens}{{\sc Ens}\xspace}

\newcommand{\hitsc}{{\sc HITS-auth}\xspace}
\newcommand{\ensc}{{\sc Ens-pseudo}\xspace}
\newcommand{\hitscs}{{\sc HITS-au}\xspace}
\newcommand{\enscs}{{\sc Ens-pse}\xspace}
\newcommand{\rand}{{\sc Random}\xspace}
\newcommand{\ifor}{{\sc iForest-r}\xspace}

\newcommand{\rnd}{{\sc Rnd}\xspace}
\newcommand{\ifr}{{\sc iF}\xspace}
\newcommand{\best}{{\sc Best}\xspace}


\newcommand{\R}{\mathbb{R}}
\newcommand{\Prob}{\mathbb{P}}

\newcommand{\mE}{\mathcal{E}}
\newcommand{\mS}{\mathcal{S}}
\newcommand{\mM}{\mathcal{M}}
\newcommand{\mD}{\mathcal{D}}
\newcommand{\mV}{\mathcal{V}}
\newcommand{\mL}{\mathcal{L}}
\newcommand{\mW}{\mathcal{W}}

\newcommand{\bX}{\bold{X}}
\newcommand{\bx}{\bold{x}}
\newcommand{\bs}{\bold{s}}
\newcommand{\bbeta}{\boldsymbol{\beta}}
\newcommand{\bmu}{\boldsymbol{\mu}}
\newcommand{\bchi}{\boldsymbol{\chi}}
\newcommand{\bPhi}{\boldsymbol{\Phi}}

\newcommand{\reminder}[1]{{\textsf{\textcolor{red}{[TODO: #1]}}}}
\newcommand{\note}[1]{{\textsf{\textcolor{blue}{[#1]}}}}
\newcommand{\summary}[1]{{\textsf{\textcolor{red}{#1}}}}

\newcommand{\hide}[1]{}

\newtheorem{problem}{Problem}



\printAffiliationsAndNotice{\icmlEqualContribution} 

\begin{abstract}

Given an unsupervised outlier detection task, how should one select a detection algorithm as well as its hyperparameters (jointly called a model)? Unsupervised model selection is notoriously difficult, in the absence of hold-out validation data with ground-truth labels. Therefore, the problem is vastly understudied. In this work, we study the feasibility of employing internal model evaluation strategies for selecting a model for outlier detection. These so-called internal strategies solely rely on the input data (without labels) and the output (outlier scores) of the candidate models. We setup (and open-source) a large testbed with 39 detection tasks and 297 candidate models comprised of 8 detectors and various hyperparameter configurations. We evaluate 7 different strategies on their ability to discriminate between models w.r.t. detection performance, without using any labels. Our study reveals room for progress---we find that none would be practically useful, as they select models only comparable to a state-of-the-art detector (with random configuration).

\end{abstract}

\vspace{-0.3in}
\section{Introduction}
\label{sec:intro}
Model selection aims to select a model from a set of candidate models for a task, given data. We consider the model selection problem for the \textit{unsupervised} outlier detection (UOD) task.
Specifically,
given
a dataset for UOD, how can we identify -- \textit{without using any labels} -- which
outlier model (a detection algorithm and the value(s) of its hyperparameter(s)) performs better than the others on the input dataset?
Importantly, note that as the outlier detection task is unsupervised, so is the model selection task. That is, an outlier model is to be selected without being able to validate any candidate models on hold-out labeled data.

The notion of a universally ``best'' outlier model does not exist; rather the best-performing model depends on the given data. On the other hand,
 model selection task is a nontrivial one, provided 
there are numerous outlier detection algorithms based on a variety of approaches; distance-based \cite{ramaswamy2000efficient,journals/vldb/KnorrNT00},   density-based \cite{breunig2000lof,tang2002enhancing,goldstein2012histogram}, 
 angle-based \cite{kriegel2008angle}, ensemble-based \cite{liu2008isolation,pevny2016loda,CharuBookEnsembles}, most recently deep neural network (NN) based \cite{conf/sdm/ChenSAT17,conf/wsdm/WangNWYL20,journals/corr/abs-2009-11732}, and so on. 
 To add to this ``choice paralysis'', most models are sensitive to their choice of hyperparameters (HPs) with significant variation in performance \cite{goldstein2016comparative}, even more so for \textit{deep} NN based outlier models that have a long list of HPs.
Unsupervised model selection will likely be an increasingly pressing problem for deep detectors, as their complexity and expressiveness grow. Recent work use some hold-out validation data for tuning such deep outlier models \cite{journals/corr/abs-2009-11732}, which however 
is not feasible for fully unsupervised settings.
 These factors make 
 outlier model selection a problem of utmost importance.

Despite its importance, the problem of unsupervised outlier model selection (UOMS hereafter) is a 
 notoriously challenging one. Mainly, the absence of validation data with labels makes the problem hard. Moreover, there does not exist a universal or well-accepted objective criterion (i.e. loss function) for outlier detection.


Perhaps due to these challenges, UOMS remains a vastly understudied area. 
Most prior work focus on designing new detection algorithms, including those for unique settings such as 
contextual \cite{conf/cikm/LiangP16,conf/pkdd/MeghanathPA18} and human-in-the-loop \cite{conf/icdm/DasWDFE16,conf/sdm/LambaA19} outlier detection. 
To our knowledge, there exist only three recent techniques specifically proposed for UOMS (in chronological order) 
\cite{MarquesCZS15,journals/corr/Goix16,JCC8455}.
In a nutshell, all of them employ \textit{internal} (i.e. unsupervised) model evaluation strategies to assess the quality of a model and its output.
However, they employ their proposed strategies to select only among 
2-3 detectors 
on 8-12 
real-world datasets. 
More problematically, they do not systematically compare to one another, nor do they use the same datasets. (See Sec. \ref{sec:related} on details of related work.)
This makes it difficult to fully understand the strengths and limitations of existing methods, and ultimately the extent to which progress has been made on this subject.

In this work, we first bring these three existing UOMS methods under one umbrella and put them to test on a large testbed. 
In addition, we apply two state-of-the-art unsupervised model selection techniques originally proposed for deep representation learning \cite{DuanMSWBLH20,lin2020infogan} to UOMS.
We also design new internal model selection methods inspired by various consensus algorithms. 
To our knowledge, this is the first work to systematically evaluate unsupervised model selection methods for outlier detection. 
We summarize the contributions and findings of this paper as follows.


\vspace{-0.05in}
\cbit

\item \textbf{Unified Comparison:~}  We identify (to our knowledge) all existing internal model evaluation strategies for UOMS. For the first time, we systematically compare them on their ability to discriminate between models w.r.t. detection performance, as well as w.r.t. running time, on the same testbed. 

\item \textbf{Large-scale Evaluation:~} Our testbed consists of 8 state-of-the-art detectors, each configured by a comprehensive list of hyperparameter settings, yielding a candidate pool of 297 models.
We perform the model selection task on 39 independent real-world datasets from two different public repositories. We compare different strategies through paired 
statistical tests to identify significant differences, if any.
We find that all three existing strategies specifically designed for UOMS are ill-suited. 
Alarmingly, none of them is significantly different from random selection (!)


\item \textbf{New UOMS Techniques:~}
All three existing methods specifically designed for UOMS are \textit{stand-alone}; evaluating each model individually, independent of others. In addition to those, we repurpose four \textit{consensus-based} algorithms from other areas for UOMS; utilizing the agreements among the models in the pool. 
We find that consensus-based methods are more competitive than stand-alone ones, and all of them achieve significantly better performance than random. However, they are not different from iForest \cite{liu2008isolation} (the best detector in our pool), thus, would not be employed (on a pool)
over training a single (iForest) model.


\item \textbf{Open-source Testbed:~}
We expect that UOMS will continue to be a pressing problem, especially with the advent of deep detection models with many hyperparameters.
Our large-scale analysis reveals that there is lots of room for progress in this field, while at the same time,
sheds light onto the strengths and limitations of different approaches that motivate various future directions. 
To 
facilitate progress on this important problem, we open-source all datasets, our trained model pool, and  
implementations of UOMS methods studied in this work at {{\url{http://bit.ly/UOMSCODE}}}.
\ceit

\section{Preliminaries \& The Problem}
\label{sec:problem}
Let $\mM = \{M_i\}_{i=1}^N$ denote a given pool of $N$ candidate models.
Here each model $M_i$ is a $\{${\tt detector}, {\tt HPconfiguration}$\}$ pair; where {\tt detector} is a certain outlier detection algorithm (e.g. LOF \cite{conf/sigmod/BreunigKNS00}) and {\tt HPconfiguration} is a certain setting of its hyperparameter(s) (e.g. for LOF, value of {\tt n\_neighbors}: number of nearest neighbors to consider, and function of choice for {\tt distance} computation). 

\begin{table*}[!t]
	\vspace{-0.1in}
\footnotesize
\centering
	\caption{Outlier Detection Models; see hyperparameter definitions from PyOD \cite{zhao2019pyod}} 
	   \scalebox{0.8}{
	\begin{tabular}{lll|r} 
	    \toprule
		\textbf{Detection algorithm} &
		\textbf{Hyperparameter 1} & \textbf{Hyperparameter 2} & \textbf{Total}\\
		\midrule
		
		   LOF \cite{breunig2000lof}                              & n\_neighbors: $[1, 5 ,10, 15, 20, 25, 50, 60, 70, 80, 90, 100]$   & distance: ['manhattan', 'euclidean', 'minkowski'] & 36\\

		    kNN \cite{ramaswamy2000efficient}                      & n\_neighbors: $[1, 5 ,10, 15, 20, 25, 50, 60, 70, 80, 90, 100]$   & method: ['largest', 'mean', 'median'] & 36\\

   		    OCSVM \cite{scholkopf2001estimating}                   & nu (train error tol): $[0.1, 0.2, 0.3, 0.4, 0.5, 0.6, 0.7, 0.8, 0.9]$    & kernel: ['linear', 'poly', 'rbf', 'sigmoid']  & 36\\

	COF \cite{tang2002enhancing}                           & n\_neighbors: $[3, 5, 10, 15, 20, 25, 50]$   & N/A & 7\\

		  ABOD \cite{kriegel2008angle}                           & n\_neighbors: $[3, 5, 10, 15, 20, 25, 50]$   & N/A  & 7\\

         iForest \cite{liu2008isolation}	    	         	& n\_estimators: $[10, 20, 30, 40, 50, 75, 100, 150, 200]$  & max\_features: $[0.1, 0.2, 0.3, 0.4, 0.5, 0.6, 0.7, 0.8, 0.9]$  & 81\\
   
         HBOS \cite{goldstein2012histogram}                     & n\_histograms: $[5, 10, 20, 30, 40, 50, 75, 100]$  & tolerance: $[0.1, 0.2, 0.3, 0.4, 0.5]$   & 40\\

		LODA \cite{pevny2016loda}                              & n\_bins: $[10, 20, 30, 40, 50, 75, 100, 150, 200]$        & n\_random\_cuts: $[5, 10, 15, 20, 25, 30]$ & 54\\

		\midrule
		&&& \textbf{297} \\ 
	\end{tabular}}
	\label{table:models_long} 
	\vspace{-0.25in}
\end{table*}

In this study, $\mM$ is composed by pairing 8 popular outlier detection algorithms to distinct hyperparameter choices, comprising a total of $N=297$ models, as listed in 
Table \ref{table:models_long}.
All models are trained based on the Python Outlier Detection Toolbox (PyOD)\footnote{\url{https://github.com/yzhao062/pyod}} on each dataset.

Let $\mD=\{D_t\}_{t=1}^T$ denote the set of outlier detection datasets (i.e. tasks), where $D_t=\{\bx_j^{(t)}\}_{j=1}^{n_t}$, $n_t=|D_t|$  is the number of samples and $o_t$ is the true number of ground-truth outliers  in $D_t$.
We denote by $\bs_i^{(t)} \in \mathbb{R}^{n_t}$ the list of outlier scores output by model $M_i$ when employed (i.e. trained\footnote{Note that as we consider unsupervised outlier detection, model ``training'' does not involve any ground-truth labels.}) on $D_t$, and $s_{ij}^{(t)} \in \mathbb{R}$ to depict individual sample $j$'s score. We omit the superscript when it is clear from context. W.l.o.g. the higher the $s_{ij}$ is, the more anomalous is $j$ w.r.t. $M_i$.

\vspace{-0.1in}
\begin{problem}[UOMS] The model selection problem for unsupervised outlier detection can be stated as follows.
	\begin{tabular}{rl}
	{Given} & \text{\em an unsupervised detection task}  $D=\{\bx_j\}_{j=1}^{n}$, \\
	& \text{\em all models in} $\mM$ \text{\em trained on} $D$ \\
	& \text{\em with corresponding output scores} $\{\bs_i\}_{i=1}^N\;$\textit{\em ;} \\	
	Select & \text{\em a model} $M'\in \mM$, \\	
	such that & $\bs'$ \text{\em yields good detection performance.} 
	\end{tabular}
\end{problem}
Note that the detection performance is to be quantified \textit{post} model selection, where ground-truth labels are used only for evaluation (and \textbf{not} for model training or model selection).

In this work, we study 7 different families of internal strategies (See Table \ref{table:overview}): (1) three techniques that were proposed to directly address the UOMS problem, (2)  two unsupervised model selection techniques adopted from deep learning, and (3) two others that are not originally designed for model selection that we adapt to UOMS. 
To compare their effectiveness systematically, we construct a large testbed of $T=39$ real-world outlier detection datasets from two different repositories (See Supp. \ref{ssec:data}). That is, we perform UOMS using each technique 39 times, to select one model from the pool of 297. Given that the datasets are independent, a large testbed enables
paired statistical tests that
 conclusively identify significant differences between these techniques as well as those and various baselines.

\section{Internal Model Evaluation Strategies}
\label{sec:measures}

Internal strategies evaluate the goodness of a model without using any external information (especially ground truth labels). The internal information being used is solely limited to ($i$) the input samples (feature values only), ($ii$) the trained models in the candidate pool and the outlier scores as output by these trained models. 


The common thread among all internal model evaluation strategies in this study is an estimated \textit{internal measure} of ``model goodness''. The model selection task is then addressed by top-1 selection: i.e. picking the model with the highest value of the respective measure.

We categorize the 7 strategies we studied into two,
depending on how they estimate their internal measure: 
(1) \textbf{stand-alone} and (2) \textbf{consensus-based}.
Stand-alone strategies solely rely on each model and its output individually, independent of other models.
All three existing methods proposed specifically for UOMS fall into this category.
On the other hand, consensus-based strategies leverage agreement between the models in the pool and hence utilize candidate models collectively. 
Four strategies we adopt and adapt\footnote{We \textit{adopt} two  strategies originally proposed for unsupervised model selection for deep representation learning ``{as is}'', and \textit{adapt} two techniques (from information retrieval and ensemble learning) by repurposing them to UOMS problem  with small modifications.}
 from other areas all fall into this latter category. 

In the following we provide a short description of each strategy (and refer to the original articles for full details). We also remark on the computational complexity of some methods as they demand considerable running time. Ideal is to have a lightweight and effective selection method with low overhead incurred on top of model training.
In the experiments, we compare these methods w.r.t. their selection performance as well as running time.


\subsection{Stand-alone internal evaluation (Existing)}

\subsubsection{IREOS}  
\label{sssec:ireos}

The first known
index proposed for the internal evaluation of outlier detection results is called
Internal, Relative Evaluation of Outlier Solutions (\ireos) \cite{MarquesCZS15}.
While their initial index is designed only for binary solutions (referred to as ``top-n'' detection),
their recent work \cite{MarquesCSZ20} generalized to numeric outlier scorings, which is the setting considered in this study.

Their intuition is that an outlier should be more easily separated (discriminated) from other samples than an inlier.
Then, a model is ``good'' the more it identifies as outlier those samples with a large degree of separability. They propose to assess the separability of each individual sample using a  maximum-margin classifier (and specifically use nonlinear SVMs).\footnote{Note that collective outliers (forming micro-clusters, or clumps) do not have high separability. \ireos accounts for this effectively, provided a user-specified {\tt clump\_size}. For details, we refer to the original articles. } The \ireos score of a model $M_i$ on a given dataset is computed as 
\beq
\small
\label{eq:ireos}
\text{\ireos}(\bs_i) = \frac{1}{n_{\gamma}} \sum_{l=1}^{n_{\gamma}} \frac{\sum_{j=1}^n p(\bx_j,\gamma_l) w_{ij}}{\sum_{j=1}^n w_{ij}}
\eeq
where $p(\bx_j,\gamma_l)$ is the separability of sample $j$ as estimated by a nonlinear SVM with kernel bandwidth (a hyper-parameter) $\gamma_l$, and $n_{\gamma}$ is the number of different bandwidth values used from the interval $[0,\gamma_{\max}]$.\footnote{They use heuristics to automatically set $\gamma_{\max}$ in their code.} They convert outlier scores $\{s_{ij}\}_{j=1}^n$ to probability weights $\{w_{ij}\}_{j=1}^n$ using the approach by \cite{conf/sdm/KriegelKSZ11} to push inlier scores toward zero so that they do not in aggregate dominate the weighted sum. 
Intuitively, \ireos tends to give high scores to those models whose outlier scores correlate well with the separability scores by a nonlinear SVM. 

Computationally, \ireos is quite demanding as it requires training of a nonlinear classifier {\em per sample}.
Their source code\footref{fnlabel} provides ways to approximate \ireos scores, mainly  estimating separability via nearest neighbor distances, which however are also expensive to compute.



\subsubsection{Mass-Volume (MV) and Excess-Mass (EM)} 
\label{sssec:goix}

\cite{journals/corr/Goix16} proposed using statistical tools, namely \mv and \emmes curves, to measure the quality of a scoring function.
Formally, a scoring function $s: \R^d \mapsto \R_{+}$ is any measurable function integrable w.r.t. the Lebesgue measure $\text{Leb}(\cdot)$, whose level sets are  estimates of the level sets of the density.
Outliers are assumed to occur in the tail of the score distribution as produced by a scoring function,
where the \textit{lower} $s(\bx)$ is, the more abnormal is $\bx$.

Given a scoring function $s(\cdot)$ (in our context, an outlier model), the \mv measure is defined as follows. 

\vspace{-0.35in}
\beq
\label{eq:mv}
\hspace{-0.0475in} \widehat{MV}_s(\alpha) = \inf_{u\geq 0} \; \text{Leb}(s\geq u) \; s.t. \;
\Prob_n(s(\bX) \geq u) \geq \alpha 
\eeq
\vspace{-0.25in}

\noindent
where $\alpha \in (0,1)$, and $\Prob_n$ is the empirical distribution; $\Prob_n(s\geq v) = \frac{1}{n} \sum_{j=1}^n \mathbbm{1}_{s(\bx_j)>v}$.

For univariate real numbers, $\text{Leb}(\cdot)$ measures the length of the given interval.
Let $s_{\max}$ denote the largest score produced by $s(\cdot)$. Then, empirically $\text{Leb}(s\geq u)$ is equal to the length $|s_{\max}-u|$. Given $\alpha$, the $u$ that minimizes the Lebesgue measure $\text{Leb}(s\geq u)$
in Eq. \eqref{eq:mv} would be equal to the outlier score at the $(1$$-$$\alpha)$-\textit{th} quantile, i.e. $u=CCDF_s^{-1}(\alpha)$.
Then,  $|s_{\max}-u|$ would give the length of the range of scores for $\alpha$ fraction of the samples with score larger than $u$. In their work, they consider $\alpha \in (0.9,0.999)$.\footnote{Assuming fraction of outliers is bounded to 10\% maximum.}$^{,}$\footnote{Area under the MV-curve is estimated as the sum of empirical MV values by Eq. \eqref{eq:mv} for discretized values of $\alpha$ in this range.} 
As they assume a \textit{lower} score is more anomalous, the Lebesgue measure quantifies the length of the interval of scores for the inliers. The smaller \mv is, the better the scoring function is deemed to be.
Intuitively, then, \mv measures the clusteredness of inlier scores (or the compactness of high-density level sets).

The \emmes measure is quite similar, and is defined as
\beq
\label{eq:em}
\hspace{-0.025in} \widehat{EM}_s(t) = \sup_{u\geq 0} \;\; \Prob_n(s(\bX) \geq u) - t \text{Leb}(s\geq u) 
\eeq
\vspace{-0.25in}

\noindent
for $t>0$. Similarly, they consider $t \in [0, \widehat{EM}_s^{-1}(0.9)]$ with $\widehat{EM}_s^{-1}(0.9) := \inf\{t\geq 0, \widehat{EM}_s(t) \leq 0.9\}$.

Intuitively, \emmes would identify as small a $u$ value as possible (so as to maximize the density mass in the first term) such that the scores larger than or equal to $u$ are as clustered as possible (so as to minimize the Lebesgue measure in the second term). Again, the more clustered are the scores of the bulk of the samples (i.e. inliers), the larger \emmes gets, and the better the scoring function is deemed to be.


\subsubsection{Clustering validation metrics} 

\cite{JCC8455} point out that a drawback of \ireos, besides computational demand, is its dependence on classification -- which itself introduces a model selection problem -- since the results may depend on the selected classification algorithm and its hyper-parameter settings.\footnote{Despite citing \ireos, they do not compare in  experiments.}$^{,}$\footnote{Another paper \cite{conf/soict/NguyenNNVNH15} by the same authors proposed a classification based internal evaluation method, similar to \ireos. Their experiments show that the current internal measures do comparably well or better with less computational overhead, hence we omit \cite{conf/soict/NguyenNNVNH15} from this study.}


Their key proposal is to apply internal validation measures for clustering algorithms to outlier detection.
As the goal of clustering is to ensure samples within each cluster are similar and different from samples in other clusters, these measures are mainly based on two criteria: compactness (capturing within cluster similarity) and/or separation (reflecting inter-cluster distance).

To that end, we split the outlier scores by a given model under evaluation for dataset $D_t$ into two clusters, denoted $C_o$ and $C_i$, respectively consisting of the highest $o_t$ scores and the rest. According to those measures, an outlier model is ``good'' the more separated these two sets of scores are and/or the more clustered 
the scores within each set are.

In their study, they compared 10 different existing clustering quality measures, such as the Silhouette index \cite{Rousseeuw1987}, Xie-Beni index \cite{journals/pami/XieB91}, etc. (See others in the original article.) To give an idea, one of the well-performing ones in our experiments, namely Xie-Beni index  of a model $M_i$, denoted $\text{\xb}_i$, is defined as follows.

\vspace{-0.15in}
\beq
\small
\text{\xb}_i = \frac{\sum_{j\in C_o} d^2(s_{ij}, c_o) + \sum_{j'\in C_i} d^2(s_{ij'}, c_i) }{n_t\; d^2(c_o,c_i)}
\eeq
where $c_o=\sum_{j\in C_o} s_{ij}/o_t$ and $c_i=\sum_{j'\in C_i} s_{ij'}/(n_t-o_t)$ depict the cluster centers and $d(\cdot,\cdot)$ is the Euclidean distance. This index can be interpreted as the ratio of the intra-cluster compactness to the inter-cluster separation.

Clustering quality based measures are typically easy to compute; most of them being linear in the number of samples.

%
%
%
%

\subsection{Consensus-based internal evaluation (Repurposed)}

%

\subsubsection{\udr} 
\label{sssec:udr}
The first consensus-based approach, namely  Unsupervised Disentanglement Ranking (\udr), is adopted from deep learning and is ``the first method for unsupervised model selection for variational disentangled representation learning'' \cite{DuanMSWBLH20}. 
Each model in their case corresponds to a $\{${\tt HPconfiguration}, {\tt seed}$\}$ pair.

Reciting Tolstoy who wrote ``Happy families are all alike; every unhappy family is unhappy in its own way.'', 
their main hypothesis is that
a model with a good hyper-parameter (HP) setting will produce \textit{similar results} under different random initializations (i.e. seeds) whereas for a poor HP setting, results based on different random seeds will look arbitrarily different.

In a nutshell, 
\udr follows 4 steps: (1) Train $N=H\times S$ models, where 
 $H$ and $S$ are the number of hyperparameter settings 
 and 
 random seeds, respectively. 
(2) For each model $M_i$,  
randomly sample (without replacement) $P\leq S$ other models with the \textit{same} HP as $M_i$, 
but \textit{different} seeds. (3) Perform $P$ pairwise comparisons between $M_i$ 
and the models sampled in Step 2 for $M_i$. (4) Aggregate pairwise similarity scores (denoted $UDR_{ii'}$) as
$UDR_i = \text{median}_{i'}\;UDR_{ii'}$, for $i=1,\ldots,N$.
Finally, they pick the model (among $N$) with the largest $UDR_i$. Intuitively, \udr selects a model with an HP setting that yields {stable} or \textit{consistent} results across various seeds.

Notice that adopting \udr for the UOMS task is trivial by making the analogy between $\{${\tt HPconfiguration}, {\tt seed}$\}$ and $\{${\tt detector}, {\tt HPconfiguration}$\}$.
While trivially applied, one may question whether the implied hypothesis (that a good detector is one that has consistent results across different HP settings) holds true for outlier models, since one of the key reasons for UOMS in the first place is that most detectors are sensitive to their HP settings \cite{goldstein2016comparative}.

They key part of \udr is how pairwise model comparisons are done in Step 3.
For UOMS, we measure the output \textit{ranking similarity} of the samples by two models,
based on three well-known measures from information retrieval \cite{conf/ictir/LiomaSL17} (See Sec. \ref{ssec:setup}).



\subsubsection{\mc} 

A follow-up work to \udr proposed ModelCentrality (\mc), which is another  consensus-based strategy for what they call ``self-supervised'' model selection for disentangling GANs \cite{lin2020infogan}. 

Their premise is similar, that ``well-disentangled models should be close to the optimal model, and hence also close to each other''.
Provided the similarity $B_{ii'}$ between two models $M_i$ and $M_{i'}$ can be computed, ModelCentrality of $M_i$ is written as $\mc_i = \frac{1}{N-1} \sum_{i' \neq i} B_{ii'}$. They then select the model with the largest $\mc_i$, which coincides with the medoid in the pool of models -- hence the name \mc.

Computationally, \mc is quadratic in the number of models as it requires all pairwise comparisons.
We also experiment with a lightweight version, called \mcs, where we randomly sample $P\leq N$ models and compute $\mc_i$ of $M_i$ as the average of its similarities to $P$ models, effectively reducing its complexity down to that of \udr.

In their experiments, \cite{lin2020infogan} report that \mc outperforms \udr schemes (Sec. \ref{sssec:udr}). Our results are consistent with their finding, possibly because it is an unrealistic hypothesis for outlier models that a good model would have consistent results across HP settings.


%

\subsubsection{Model Centrality by HITS}

We can build on the idea of ModelCentrality through computing centrality in a network setting. 
Unlike \mc that 
is computed in one-shot, network centrality is \textit{recursive}---wherein a node has higher centrality the more they point to nodes that are pointed by other high-centrality ones.

One of the earliest methods for computing centrality, namely hubness $h_p$ and authority $a_p$, of pages on the Web is the HITS algorithm \cite{kleinberg1999hits}, where

\vspace{-0.3in}
\begin{align*}
	h_p &\propto \text{sum of }  a_i \text{ for all nodes } i \text{ that } p \text{ points to} \;, \; \text{and}\\
	a_p &\propto \text{sum of }  h_i \text{ for all nodes } i \text{ pointing to } p \;,
\end{align*}
\vspace{-0.3in}

\noindent
which are estimated alternatingly over iterations until convergence.
Besides ranking on the Web, HITS-like ideas have been used to estimate user trustworthiness in online rating platforms \cite{conf/icdm/WangXLY11,conf/wsdm/KumarHMKFS18}, physician authoritativeness in patient referral networks \cite{conf/sdm/MishraA17}, polarity of subjects in political networks \cite{conf/icwsm/Akoglu14}, as well as truth discovery \cite{Yin2007Truth}.

\begin{table}[t]
	\vspace{-0.1in}
\centering
	\caption{ Overview of UOMS methods in this study } 
	   \scalebox{0.665}{
	\begin{tabular}{lrrr} 
	    \toprule
		 \textbf{Method} &  \textbf{Type} & \textbf{Based on} & \textbf{Strategy} \\
		\midrule
		 \xb,\rs,$\ldots$ \cite{JCC8455} & Stand-alone & Outlier scores & Cluster quality \\
 		\hline
 		\emmes, \mv  \cite{journals/corr/Goix16} & Stand-alone& Outlier scores& Level sets\\
 		\hline
 		\ireos	\cite{MarquesCZS15} & Stand-alone & O. scores + Input & Separability\\
 		\hline
 		\udr \cite{DuanMSWBLH20}& Consensus & Outlier scores& One-shot\\
 		\mc \cite{lin2020infogan}, \mcs& Consensus & Outlier scores& One-shot\\
 			\hline
 		\hits \cite{kleinberg1999hits} & Consensus& Outlier scores& Iterative\\
 		\ens \cite{journals/sigkdd/ZimekCS13}& Consensus& Outlier scores& Iterative\\
 		\bottomrule
	\end{tabular}}
	\label{table:overview} 
	\vspace{-0.11in}
\end{table}

It is easy to adapt \hits for UOMS by constructing a complete bipartite network between the $N$ models and $n_t$ samples in a given dataset $D_t$. Then, the models can be evaluated by their hubness centralities.
The analogous interpretation is that a
a sample has higher authority (outlierness), the more trusted models (with high hubness) point to it (with large outlierness score, i.e. large edge weight). 
Then, a model is more central or trusted, the more it points (with large outlierness score) to samples with high authority. 

Note that a by-product of this strategy is a consensus-based ranking of the samples based on authority scores (i.e. centrality-based outlierness)  upon convergence. We compare this (aggregate) ranking, called \hitsc, against selecting a (single) model by hubness in the experiments.  

%

\subsubsection{Unsupervised outlier model ensembling}

HITS has a built-in advantage that is the iterative refinement of model trustworthiness.
Specifically,
given the trustworthiness of models, outlier scores can be better estimated by a {trustworthiness-weighted} aggregation of scores across models.
Then, given those refined outlier scores, model trustworthiness can also be better estimated; where 
the more similar their output is to the updated scores, the more a model is deemed trustworthy.

In this part we build on another iterative scheme, originally designed for unsupervised selective outlier model ensembling \cite{journals/sigkdd/ZimekCS13,journals/tkdd/RayanaA16}.
The idea is to infer reliable ``pseudo ground truth'' outlier scores via aggregating the output of a carefully-selected subset of trustworthy models. The ensemble is constructed bottom-up in a greedy iterative fashion, as outlined in Alg. \ref{alg:select}. 

Similar to HITS, the ``pseudo ground truth'' and model trustworthiness are estimated alternatingly. The latter is computed as the ranking based similarity of a model's output to the ``pseudo ground truth'' (i.e. $target$ in Alg. \ref{alg:select}) at a given iteration. We adapt this framework to UOMS by using these similarities at convergence to evaluate the models. We call this strategy \ens. 
In experiments, we also compare the (aggregate) ranking by the ensemble (based on $target$), called \ensc,
to
selecting a (single) model (with highest similarity to $target$).


To wrap up,  we give a summary of the 7 families of UOMS techniques as described in this section in Table \ref{table:overview}.

	\vspace{-0.05in}
\section{Related Work}
\label{sec:related}
Related work on UOMS is slim, with only a few existing techniques that we already covered in the previous section.
Here
we provide a brief critique and comparison.

\begin{algorithm}[t]
	\caption{Ensemble-based Internal Model Evaluation}
	\label{alg:select}
	\begin{algorithmic}[1]
		\small{
		\REQUIRE set of outlier scores from all models, $\{\bs_i\}_{i=1}^N$ 
		\ENSURE internal scores for all models
		\STATE $\mS := \emptyset\;$, $\mE := \emptyset$, $C := 0$  
		\FOR[convert scores to inverse rank]{$i=1,\ldots, N$}
		\STATE $\mS := \mS \cup \{1/\text{rank}(s_{ij}) \}_{j=1}^n$
		\ENDFOR
		\STATE $target := \text{avg}(\mS)$ \COMMENT{initial pseudo ground truth scores} 
		%
		\REPEAT
		\STATE sort $\mS$ by rank $corr$elation to $target$ in desc. order
		\STATE $\{m, \text{corr}_m\} := \text{fetchFirst}(\mS)$
		\IF{$corr(\text{avg}(\mE\cup m), target) \times |E| \geq {C} $}
		\STATE $\mE := \mE \cup m$, \quad $C+$$=\text{corr}_m$
		\STATE $target := \text{avg}(\mE)$ \COMMENT{pseudo ground truth by $\mE$}
		\ENDIF
		\UNTIL{$\{$$\mS = \emptyset$ or $\mE$ is not updated$\}$}
		\STATE {\bf return} rank $corr$elation of $\bs_i$ to $target$, $i=1,\ldots, N$
	}
	\end{algorithmic}
\end{algorithm}
\setlength{\textfloatsep}{0.15in}



{\bf Existing methods for UOMS:~} Cluster quality based measures \cite{JCC8455} and statistical mass based \emmes/\mv methods \cite{journals/corr/Goix16} rely only on output scores. In contrast \ireos  \cite{MarquesCZS15,MarquesCSZ20} uses more information, that is both outlier scores and the original input samples (See Eq. \eqref{eq:ireos}).
Verifying that outlier scores align (correlate) with the separability of samples in the feature space is potentially less error-prone than simply looking at whether outlier/inlier scores are well clustered or separated -- e.g., a model that outputs a $\{0,1\}$ score per point at random would be considered a good model by the latter.
The trade-off is the computational overhead for quantifying separability per sample.


In their work, \ireos is employed for UOMS using only 2 detectors (LOF \cite{breunig2000lof} and kNN \cite{ramaswamy2000efficient}), each with 17 different HP configurations (for a total of 34 models) on 11 datasets. Being the seminal work, there is no comparison to any other techniques (existing or adapted).
\cite{JCC8455} acknowledge \ireos and criticize its computational demand, without any comparison.
They also do not perform any UOMS in experiments, rather, they study the decay in internal measures as the ground truth ranking is contaminated via random swaps at the top based on 12 datasets. Finally, \cite{journals/corr/Goix16} performs UOMS using only and exactly 3 models (LOF, iForest \cite{liu2008isolation}, OCSVM \cite{scholkopf2001estimating}), each with a single (unspecified) HP configuration, on 8 datasets.
None of these three compares to any other in their work. Moreover, because
the datasets, experimental design, and the model pool specified by each work is different, it is not possible to do any direct comparison.
In this work,  we do a systematic comparison for the first time, using a much larger testbed (8 detectors, 297 models, 39 datasets) than originally considered by any prior work.

{\bf Repurposed methods for UOMS:} All three existing methods for UOMS are stand-alone, evaluating a model independent  from others. 
Having trained all models among which to select from, it is reasonable to take advantage of the similarities/agreement among them. To this end, we have repurposed methods from unsupervised representation learning \cite{DuanMSWBLH20,lin2020infogan}, network centrality \cite{kleinberg1999hits}, and unsupervised ensemble learning \cite{journals/sigkdd/ZimekCS13,journals/tkdd/RayanaA16} all of which are based on the ``collective intelligence'' of the models in the pool.

As we show in experiments, these strategies produce superior outcomes than existing, stand-alone methods.
As such, our study motivates and calls for the transfer of prominent ideas from other similar fields, such as truth discovery and crowdsourcing, to address the important problem of UOMS.



%
%
%

\section{Experiments}
\label{sec:experiment}

\subsection{Setup}
\label{ssec:setup}

\textbf{Datasets and Model Pool.} We already discussed the real-world datasets and candidate models of this study in Sec. \ref{sec:problem}.
As quick reference, details of our $T$$=$$39$ datasets can be found in Supp. \ref{ssec:data} and
the specifications for all $N$$=$$297$ models have been listed in Sec. \ref{sec:problem} Table \ref{table:models_long}.

\textbf{Baselines.}
We compare 
the model selected by each technique (Sec. \ref{sec:measures}) to two baselines across datasets. 

\vspace{-0.05in}
\cbit
\item \rand, whose performance is the average of all (297) models per dataset. This is equivalent to expected performance when selecting a model from the candidate pool at random. 
\item \ifor, with performance as the average of all (81) iForest models in the pool, equivalent to using iForest  \cite{liu2008isolation} (a state-of-the-art ensemble detector) with randomly chosen hyperparameters.\footnote{Family-wise performances across datasets (See Supp. \ref{ssec:family}) show that iForest is the most competitive among the 8 families of detectors used in this study, and hence the strongest baseline.}
\ceit


\textbf{Method Configurations.~} 
Due to space limit, details are given in Supp. \ref{ssec:config}.

\hide{
For clustering-quality based measures, we split into two clusters as the top $o_t$ (true number of outliers) and the rest, i.e. give those strategies the advantage of knowing $o_t$. This is to avoid the  clustering step, which requires us to pick a clustering algorithm etc., and directly focus on the measures themselves.

For \emmes and \mv\footnote{\scriptsize{\url{https://github.com/ngoix/EMMV_benchmarks}}}, we use the default values for $\alpha$ and $t$ respectively (See Sec. \ref{sssec:goix}) and set {\tt n\_generated}$=100K$, which is the number of random samples to generate for estimating the null distributions.

For \ireos, we 
use the recommended settings by the authors;\footnote{We thank Henrique Marques who helped with running their source code, \scriptsize{\url{https://github.com/homarques/ireos-extension}\label{fnlabel}}}
$\gamma_{\max}$$:=$$\text{findGammaMaxbyDistances}(\cdot)$ with {\tt sampling}$=$$100$,
{\tt tol}$=$$5\times 10^{-3}$, and {\tt clump\_size}$=$$10$. 


For \udr, \mc, and \mcs, we experiment with three different pairwise similarity measures: Spearman's $\rho$,
Kendall's $\tau$, and NDCG \cite{conf/ictir/LiomaSL17}.
For \mcs, $P=\sqrt{N}\approx 18$.

%

For \hits and \ens, we set 
edge weights 
between model $M_i$ and sample $j$ in a dataset 
as $1/r_{ij}$, where $r_{ij}$ is the position of $j$ in the rankedlist by $M_i$. Raw outlier scores are not used as they are not comparable across models. 
For comparison between selection versus consensus/ensembling, we also report the performance of the consensus outcome, called \hitsc and \ensc; as ranked (resp.)  by authority scores and by the pseudo ground truth at convergence.

}

\textbf{Performance metrics.~} We evaluate performance w.r.t. three metrics. Two are based on the ranking quality: 
Average Precision (\textbf{AP}): the area under the precision-recall curve and 
\textbf{ROC} AUC: the area under the recall-false positive rate curve. 
The third metric measures the quality at the top:
\textbf{Prec@}$\bm{k}$, precision at top $k$ where we set $k=o_t$ (i.e. true number of outliers) for each $D_t \in \mD$.
In Supp. \ref{ssec:performances} we show that performances vary considerably across models for most datasets, justifying the importance of model selection.

Due to space limit, all results in this section are w.r.t. AP. 
Corresponding results for other metrics are similar, all of which are provided  in Supp. \ref{ssec:other}.

\subsection{Results}

\vspace{-0.05in}
{\bf Cluster quality based methods.~} 
We start by studying the 10 cluster quality based methods to identify those that stand out.
We report the 
$p$-values by the one-sided\footnote{Testing the hypothesis: row-method is better than col-method (against the null hypothesis stating no difference).  For reverse order, $p$-value is equal to 1 minus the reported value.} paired Wilcoxon signed rank test in Table \ref{table:pval12}.
\std is significantly worse than all other methods.
Three strategies that stand out are \rs, \ch, and \xb, which are identical; in the sense that despite differences in their values and overall ranking, they select exactly the same model on each dataset.
Importantly, while both \std and \s are significantly worse than \rand at $p=0.05$,
 none of the others is significantly different from \rand (!)
 All methods (including \xb, \rs, and \ch) are significantly worse than \ifor.

%
%
%
%


\begin{table}[!h]
	\vskip -0.2in
	\footnotesize
\centering
	\caption{Comparison of cluster quality based methods and baselines
		 by one-sided paired Wilcoxon signed rank test.
		$p$-values \textbf{bolded} (\underline{underlined}) highlight the cases where row-method  is significantly \textbf{better} (\underline{worse}) than col-method at $p$$\leq$$0.05$.
	} 
	\vspace{0.05in}
	   \scalebox{0.7}{
	   		\begin{tabular}{l|lllllll|ll}
	   			\toprule
	   		&	\std & \h &  \s &\iind & \db &  \sd & \dunn & \rnd & \ifr \\
	   		\midrule
	   		\xb,\rs,\ch & \textBF{0.004} & 0.240  &  \textBF{0.038} & 0.212 & 0.370    & 0.127 & 0.357 & 0.500   & \underline{0.981} \\
	   		\std &      & \underline{0.997} & \underline{0.961} & \underline{0.997} & \underline{0.982} &  \underline{0.967} & \underline{0.999} & \underline{1.000}     & \underline{1.000}     \\
	   		\h &      &      &  0.373 & 0.500   & 0.725 &  0.379 & 0.675 & 0.849 & \underline{0.996} \\
	   		\s &      &            &     & 0.627 & 0.949 & 0.557 & 0.881 & \underline{0.953} & \underline{0.999} \\
	   		\iind &      &           &      &      & 0.730  & 0.384 & 0.742 & 0.882 & \underline{0.997} \\
	   		\db &      &           &      &      &        & 0.307 & 0.647 & 0.522 & \underline{0.982} \\
	   		\sd &   &           &      &      &      &      & 0.823 & 0.910  & \underline{0.995} \\
	   		\dunn &      &           &      &      &      &     &      & 0.572 & \underline{0.990}  \\
	   		\rnd &      &            &      &      &      &      &      &      & \underline{1.000}    
%
	   		\end{tabular}
}
	\label{table:pval12} 
\vspace{-0.15in}
\end{table}

These findings suggest that cluster quality based internal evaluation methods would not be useful for UOMS.

{\bf Other stand-alone methods.~} As discussed in Sec. \ref{sssec:goix},
\emmes and \mv quantify (roughly) the clusteredness of the inlier scores. Therefore, they are conceptually 
similar to the clustering quality based methods.
Our findings confirm this intuition. As shown in Table \ref{table:emmv}, there is no significant difference between \emmes/\mv and \xb/\rs/\ch or \rand. Both of them are also significantly worse than \ifor. Thus, they do not prove useful for UOMS.
Findings are similar for \ireos; despite using more information (input samples besides scores, see Eq. \eqref{eq:ireos}) 
and computational cost, it is only comparable to \rand.


\begin{table}[!h]
	\vspace{-0.125in}	
	\footnotesize
	\centering
	\caption{Comparison of stand-alone methods and baselines.} 
	\scalebox{0.9}{
		\begin{tabular}{l|lll|ll}
			\toprule
			&  \emmes	& \mv & \ireos & \rnd & \ifr \\
			\midrule
			\xb,\rs,\ch & 0.533 & 0.500 & 0.862 & 0.500   & \underline{0.981}  \\	
			\emmes & & 0.079 & 0.642 & 0.539 &  \underline{0.979} \\
			\mv& & & 0.716 & 0.687 &  \underline{0.994} \\
			\ireos & & &  & 0.303 &  0.908  \\
			\bottomrule
		\end{tabular}
	}
	\label{table:emmv} 
	\vspace{-0.05in}	
\end{table}

We provide an additional viewpoint by identifying the $q$\textit{-th best} model per dataset where there exists no significant difference between the performance of the $q$-th best model and that selected by a given UOMS strategy across datasets.
We report the \textit{smallest} $q$ for which one-sided Wilcoxon signed rank test yields $p$$>$$0.05$ in
Table \ref{table:q}. 
A method with smaller $q$ is better; the interpretation being that it could select, from a pool of 297, the model that
is as good as the $q$-{th} best model per dataset. 
Stand-alone methods do not fare well against \ifor which is comparable to the $84$-{th} 
best model. 

\begin{table}[!h]
	\vskip -0.1in
\centering
	\caption{\textbf{Summary of results}: $p$-values by one-sided paired Wilcoxon signed rank test comparing UOMS methods to the baselines, smallest $q$-th best model with no significant difference, and mean/standard deviation AP across datasets. 
	} 
	   \scalebox{0.75}{
	\begin{tabular}{cl|cc|r|cc} 
	    \toprule
		& \textbf{Method} &  \rand & \ifor & \qap & mean AP & std AP \\ 
		\midrule
		\multirow{3}{*}{\rotatebox[origin=c]{90}{\parbox[c]{1.1cm}{\centering S-alone}}}
		& \xb,\rs,\ch    & 0.500 & \underline{0.981} &	127 & 0.354 & 0.298 \\ 
 		\cline{2-7}
 		&\emmes  &  0.539 &  \underline{0.979} & 115 & 0.322 & 0.265 \\ 
 		\cline{2-7}
 		&\ireos	 & 0.303  & 0.908  & 99 & 0.335 & 0.261\\ 
 		 
 		\midrule
 		\multirow{11}{*}{\rotatebox[origin=c]{90}{\parbox[c]{4cm}{\centering Consensus-based}}}
	&	\udr-${\rho}$ & \textBF{0.012} & 0.905 & {104} & {0.383}& {0.283}  \\
	&	\udr-$\tau$ & \textBF{0.019} & \underline{0.952} & 109	& 0.379	& 0.282	\\
	&	\udr-$NDCG$ & \textBF{0.004} & {0.825} &  93 & 0.384 & 0.270	\\
	\cline{2-7}
	&	\mc-${\rho}$& \textBF{0.000} & 0.217 & 89	& 0.395	& 0.289	\\
	&	\mc-${\tau}$& \textBF{0.002} & 0.062	& {81}	& {0.396}	& {0.297} \\
	&	\mc-${NDCG}$& \textBF{0.000} & {0.182}	& 82 & 0.404 & 0.291
	\\
		\cline{2-7}
	&	\mcs-${\rho}$& \textBF{0.007} & 0.706 	& 108 & 0.385 & 0.289 \\	
	&	\mcs-${\tau}$& \textBF{0.001} & 0.599 & 90 & 0.397 & 0.305 \\
	&	\mcs-${NDCG}$&  \textBF{0.001} & {0.205}& 83 & 0.391 & 0.285 \\
 				
 			\cline{2-7}
 		&\hits& \textBF{0.000} & 0.494 & 95 & 0.397& 0.299 \\
 		&\ens& \textBF{0.002} & 0.730 & 81 &0.371& 0.282 \\
 		 	\midrule
 		 	\multirow{2}{*}{\rotatebox[origin=c]{90}{\parbox[c]{0.6cm}{\centering Agg.}}}	
 		 	&\hitsc& \textBF{0.000} & 0.577 & 94& 0.401& 0.286\\ 
 		 &\ensc& \textBF{0.001} & 0.422 & 79 & 0.373 & 0.282	 \\
 				\midrule	
 			\multirow{2}{*}{\rotatebox[origin=c]{90}{\parbox[c]{1cm}{\centering Base.}}}	
 		&\rand &\cdash 	& \underline{1.000} & 144 & 0.342 & 0.234\\ 
 		&\ifor 	&\cdash & \cdash & 84 & 0.399 & 0.300 \\ 
 		\bottomrule
	\end{tabular}}
	\label{table:q} 
	\vskip -0.1in
\end{table}

{\bf Consensus-based methods.~}
We first study one-shot methods \udr, \mc, and \mcs based on different similarity measures.
As shown in Table \ref{table:q}, all versions provide similar results, which
are significantly better than \rand, and not different from \ifor.
We note that the faster, sampling-based \mcs achieves similar performance to \mc and can be used as a practical alternative. 

Iterative methods \hits and \ens produce similar results to these simple one-shot methods, despite aiming to refine estimates of model trustworthiness  over iterations.
Again, as shown in Table \ref{table:q}, they significantly outperform \rand and are comparable to \ifor.
The same holds true for their respective consensus scores, \hitsc and \ensc, where 
model aggregation provides no significant advantage over
selecting the best (single) model. 

Table \ref{table:consensus} shows a pairwise comparison of the consensus-based methods by one-sided Wilcoxon signed rank test, confirming mostly no significant difference between them.


\begin{table}[!h]
\vspace{-0.15in}	
	\footnotesize
	\centering
	\caption{Comparison of consensus-based methods (\udr, \mc, \mcs are based on ${NDCG}$). } 
	\scalebox{0.9}{
		\begin{tabular}{l|llll}
			\toprule
			&   \mc  & \mcs & \hits & \ens \\ 
			\midrule
			\udr & 0.810 & 0.364 & 0.739 & 0.400 \\ 
			\mc  & & 0.551 &  \textBF{0.039} &  0.116\\ 
			\mcs  & & & 0.296 & 0.369 \\ 
			\hits  & & & & 0.753 \\ 
			\bottomrule
		\end{tabular}
	}
	\label{table:consensus} 
\end{table}

{\bf Running time analysis.~}
In Fig. \ref{fig:runtime} we present for each method the running times on all datasets.\footnote{On an Intel Xeon E7 4830 v3 @ 2.1Ghz with 1TB RAM} 
\ireos and \emmes/\mv are both computationally demanding, while ineffective. In fact,
 \ireos takes more than 16 days (!) on the largest dataset (ALOI), due to kernel SVM training {\em per} sample. \mc is the next most expensive method, which is quadratic in the number of models, but still takes less than 1 hr on ALOI.
 In short, \mcs, \ens, and especially \hits prove to be both competitive as well as fast UOMS methods, completing within 10 minutes on our testbed. 



\begin{figure}[!h]
	\vskip -0.05in
	\begin{center}
			\centerline{\includegraphics[width=1\columnwidth]{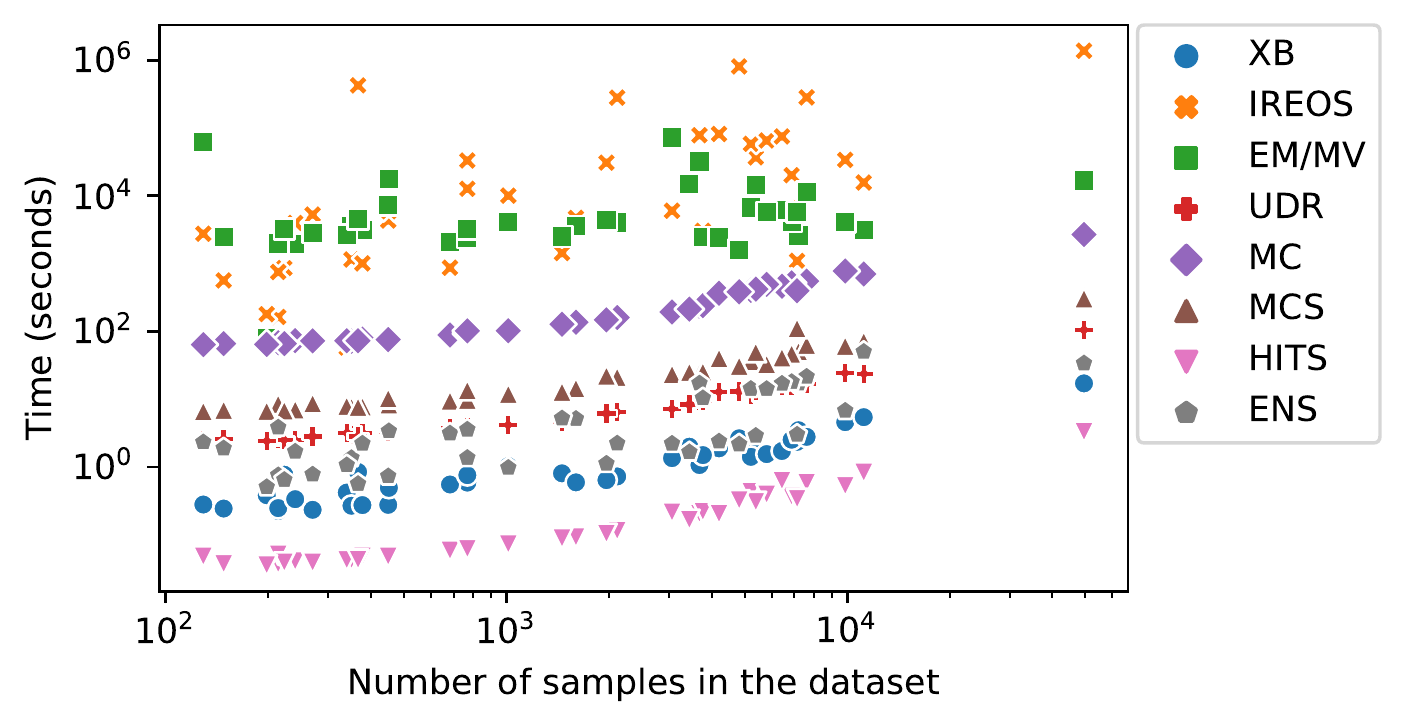}}
		\vskip -0.175in
		\caption{Run time comparison of UOMS methods.
		}
		\label{fig:runtime}
	\end{center}
	\vskip -0.25in
\end{figure}

\subsection{Discussion of the Results}

Key take-aways from our study are two: (1) \textit{None of the existing (stand-alone) UOMS methods is significantly different from random model selection} (!), and with the exception of \ireos, all are significantly worse than iForest (with random hyperparameter configuration). The slight advantage of \ireos can be attributed to it utilizing input features in addition to model outlier scores, at the expense of significant running time.
(2) \textit{All consensus-based methods that we repurposed for UOMS are significantly better than random selection, but not different from iForest}. 

Fig. \ref{fig:diff} 
illustrates these take-aways where we show, via boxplots, the distribution of the performance difference between the model selected by each UOMS method and \ifor across datasets. Consensus-based methods select models at best as good as \ifor, where the AP difference concentrates around zero, whereas others are inferior.

\begin{figure}[!t]
 \vskip -0.05in
	\begin{center}
	\hspace{-0.2in}	\centerline{\includegraphics[width=0.9\columnwidth,height=1.35in]{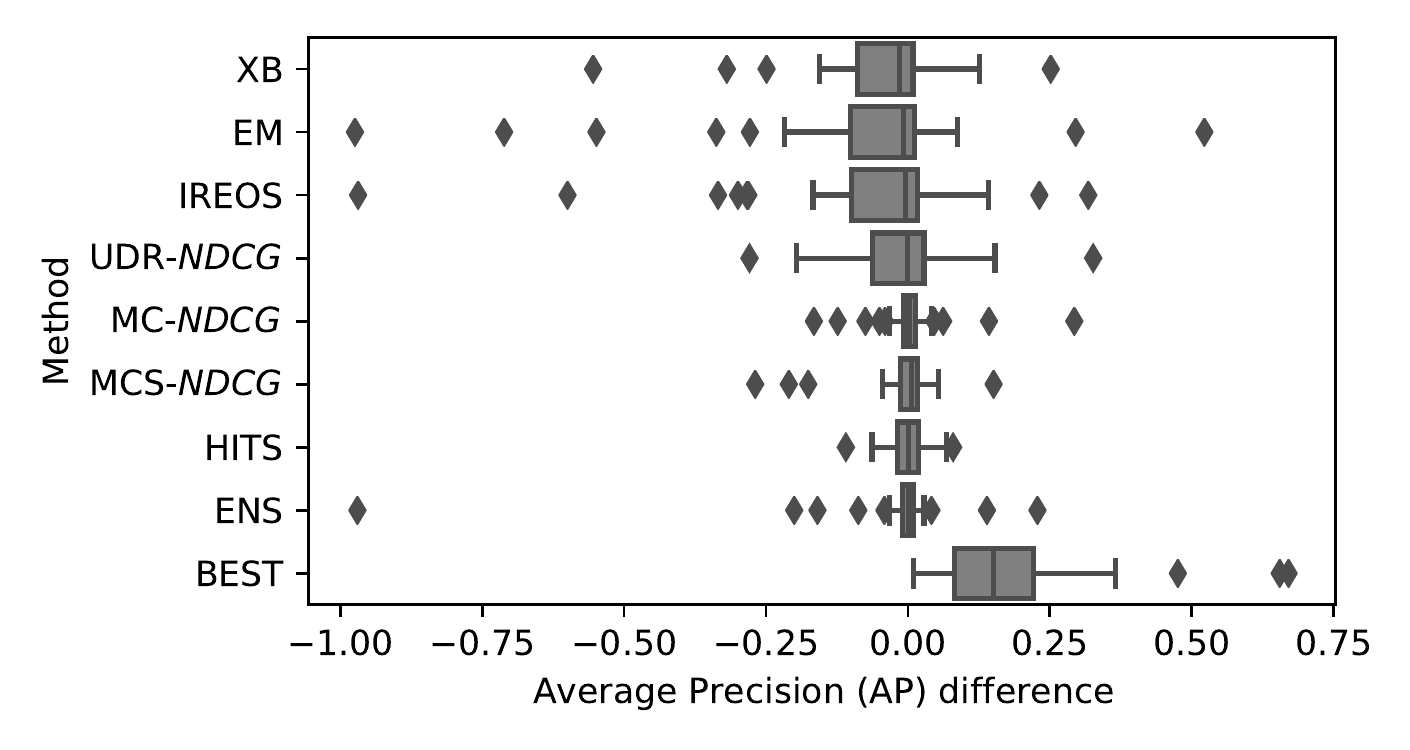}}
		\vskip -0.175in
		\caption{Distribution across datasets of performance difference: AP of selected model (by each UOMS method studied) minus that of \ifor. Stand-alone methods and \udr are subpar, whereas other consensus-based method differences concentrate around zero (indicating no notable difference from \ifor). Also shown for comparison is \best model on each dataset. 
		}
		\label{fig:diff}
	\end{center}
	\vskip -0.1in
\end{figure}

These results suggest that \textbf{none of the UOMS methods we studied would be useful in practice}; because one would not first train a large \textit{pool} of models -- which would incur considerable computation -- and then run a post hoc UOMS method to select a model,  only to achieve comparable performance to a \textit{single} iForest model (even with random configuration) -- which, in contrast, is 
extremely fast to train as it builds randomized trees on subsamples of data.

However, this is not to conclude iForest is the best that one can hope to do. 
As given in Table \ref{table:q}, \ifor is only as good as the 84-th best model per dataset. While  it is the most competitive detector on average, other families outperform iForest on 28 out of 39 datasets in our study w.r.t. AP (See Table \ref{table:model_performance_ap} in Supp. \ref{ssec:family}, also see Tables \ref{table:model_performance_roc} and \ref{table:model_performance_prn} respectively for ROC and Prec@$k$). In Fig. \ref{fig:diff} we also show the performance difference of the 1-st \best model per dataset from \ifor. (Also see Fig.s \ref{fig:diffroc} and \ref{fig:diffprn} in Supp. \ref{ssec:other}.) One can clearly recognize that there is considerable room for progress in the area of UOMS.


\section{Conclusion}
\label{sec:conclusion}

\vspace{-0.05in}
We considered the unsupervised outlier model selection (UOMS) problem: Given an unlabeled dataset, which outlier detection algorithm and hyperparameter settings should one use?
On a large testbed comprising 297 models and 39 real-world datasets, we compared 7 different internal model evaluation strategies. 
We find that, while consensus-based approaches are more promising against stand-alone strategies, 
none of these provides significant improvement over the state-of-the-art iForest detector.
This calls for further research in this area.
A promising future direction is to employ UOMS strategies within a meta-learning framework for model selection, which can guide model search based on sequential model-based optimization \cite{jon01}.  
UOMS  also provides fertile ground for adaptation of prominent
consensus-based techniques from related areas such as truth discovery, crowdsourcing, etc.
To facilitate further progress on this critical problem, we publicly share all source code and datasets at 
~{\url{http://bit.ly/UOMSCODE}}.

\bibliography{refsMetaOD}
\bibliographystyle{icml2020}

\clearpage
\newpage

\appendix 
\section{Appendix}
\label{sec:appendix}

\subsection{Real-world Outlier Detection Datasets}
\label{ssec:data}

We build the experiments on \textbf{39 widely used outlier detection benchmark dataset}. As shown in Table \ref{table:datasets}, 21 datasets are from the \textbf{ODDS Library}\footnote{\url{http://odds.cs.stonybrook.edu}}, and the other 18 datasets are from \textbf{DAMI datasets} \cite{Campos2016}\footnote{\url{http://www.dbs.ifi.lmu.de/research/outlier-evaluation/DAMI}}.

\begin{table}[!htp]
	\centering
	\caption{Real-world dataset pool composed by ODDS library (21 datasets) and DAMI library (18 datasets).} 
	\scriptsize
	\begin{tabular}{l l | r  r r } 
		\toprule
		& \textbf{Dataset}                 & \textbf{Num Pts} & \textbf{Dim} & \textbf{\% Outlier} \\
		\midrule
		1 & annthyroid (ODDS)             & 7200         & 6            & 7.416                       \\
		2 & arrhythmia (ODDS)             & 452          & 274          & 14.601                      \\
		3 & breastw (ODDS)                & 683          & 9            & 34.992                      \\
		4 & glass (ODDS)                  & 214          & 9            & 4.205                       \\
		5 & ionosphere (ODDS)             & 351          & 33           & 35.897                      \\
		6 & letter (ODDS)                 & 1600         & 32           & 6.250                         \\
		7 & lympho (ODDS)                 & 148          & 18           & 4.054                       \\
		8 & mammography (ODDS)            & 11183        & 6            & 2.325                        \\
		9 & mnist (ODDS)                  & 7603         & 100          & 9.206                       \\
		10 & musk (ODDS)                   & 3062         & 166          & 3.167                       \\
		11 & optdigits (ODDS)              & 5216         & 64           & 2.875                       \\
		12 & pendigits (ODDS)              & 6870         & 16           & 2.270                       \\
		13 & pima (ODDS)                   & 768          & 8            & 34.895                      \\
		14 & satellite (ODDS)              & 6435         & 36           & 31.639                      \\
		15 & satimage-2 (ODDS)             & 5803         & 36           & 1.223                       \\
		16 & speech (ODDS)                 & 3686         & 400          & 1.654                       \\
		17 & thyroid (ODDS)                & 3772         & 6            & 2.465                       \\
		18 & vertebral (ODDS)              & 240          & 6            & 12.500                         \\
		19 & vowels (ODDS)                 & 1456         & 12           & 3.434                       \\
		20 & wbc (ODDS)                    & 378          & 30           & 5.555                       \\
		21 & wine (ODDS)                   & 129          & 13           & 7.751                       \\
		\midrule
		22 & Annthyroid (DAMI)             & 7129         & 21           & 7.490                       \\
		23 & Arrhythmia (DAMI)             & 450          & 259          & 45.777                      \\
		24 & Cardiotocography (DAMI)       & 2114         & 21           & 22.043                      \\
		25 & HeartDisease (DAMI)           & 270          & 13           & 44.444                      \\
		26 & InternetAds (DAMI)            & 1966         & 1555         & 18.718                      \\
		27 & PageBlocks (DAMI)             & 5393         & 10           & 9.456                       \\
		28 & Pima (DAMI)                   & 768          & 8            & 34.895                      \\
		29 & SpamBase (DAMI)               & 4207         & 57           & 39.909                      \\
		30 & Stamps (DAMI)                 & 340          & 9            & 9.117                       \\
		31 & Wilt (DAMI)                   & 4819         & 5            & 5.333                       \\
		32 & ALOI (DAMI)                   & 49534        & 27           & 3.044                       \\
		33 & Glass (DAMI)                  & 214          & 7            & 4.205                       \\
		34 & PenDigits (DAMI)              & 9868         & 16           & 0.202                       \\
		35 & Shuttle (DAMI)                & 1013         & 9            & 1.283                       \\
		36 & Waveform (DAMI)               & 3443         & 21           & 2.904                       \\
		37 & WBC (DAMI)                    & 223          & 9            & 4.484                       \\
		38 & WDBC (DAMI)                   & 367          & 30           & 2.724                       \\
		39 & WPBC (DAMI)                   & 198          & 33           & 23.737                      \\
		\bottomrule
	\end{tabular}
	\label{table:datasets} 
\end{table}

\vspace{0.5cm}
\subsection{Model Configurations}
\label{ssec:config}

For clustering-quality based measures, we split into two clusters as the top $o_t$ (true number of outliers) and the rest, i.e. give those strategies the advantage of knowing $o_t$. This is to avoid the  clustering step, which requires us to pick a clustering algorithm etc., and directly focus on the measures themselves.

For \emmes and \mv\footnote{\scriptsize{\url{https://github.com/ngoix/EMMV_benchmarks}}}, we use the default values for $\alpha$ and $t$ respectively (See Sec. \ref{sssec:goix}) and set {\tt n\_generated}$=100K$, which is the number of random samples to generate for estimating the null distributions.

For \ireos, we 
use the recommended settings by the authors;\footnote{We thank Henrique Marques who helped with running their source code, \scriptsize{\url{https://github.com/homarques/ireos-extension}\label{fnlabel}}}
$\gamma_{\max}$$:=$$\text{findGammaMaxbyDistances}(\cdot)$ with {\tt sampling}$=$$100$,
{\tt tol}$=$$5\times 10^{-3}$, and {\tt clump\_size}$=$$10$. 


For \udr, \mc, and \mcs, we experiment with three different pairwise similarity measures: Spearman's $\rho$,
Kendall's $\tau$, and NDCG \cite{conf/ictir/LiomaSL17}.
For \mcs, $P=\sqrt{N}\approx 18$.

%

For \hits and \ens, we set 
edge weights 
between model $M_i$ and sample $j$ in a dataset 
as $1/r_{ij}$, where $r_{ij}$ is the position of $j$ in the rankedlist by $M_i$. Raw outlier scores are not used as they are not comparable across models. 
For comparison between selection versus consensus/ensembling, we also report the performance of the consensus outcome, called \hitsc and \ensc; as ranked (resp.)  by authority scores and by the pseudo ground truth at convergence.


\vspace{0.5cm}
\subsection{Family-wise Model Performances}
\label{ssec:family}

In this study we use \textbf{8 different families of outlier detection algorithms}, 
namely; {LODA}, {ABOD}, {iForest}, {kNN},  {LOF}, {HBOS}, {OCSVM}, and {COF}.
We build a total of \textbf{297 detection models} based on various hyperparameter (HP) configurations of these algorithms, 
as listed in Table \ref{table:models_long}.

Tables \ref{table:model_performance_ap}, \ref{table:model_performance_roc}, and \ref{table:model_performance_prn} (resp. for AP, ROC AUC, and Prec@$k$) show the family-wise average performance of each detection algorithm (averaged over within-family models with different HP settings) on each dataset, as well as mean and standard deviation across datasets.

These show \textbf{{iForest} to be the most competitive detector}, which we compare to as a baseline to study \textit{whether unsupervised model selection outperforms always using the same (state-of-the-art) detector}.

\vspace{0.5cm}
\subsection{Model Performances on Individual Datasets}
\label{ssec:performances}

Figures \ref{fig:APmodel}, \ref{fig:ROCmodel}, and \ref{fig:PRNmodel} (resp. for AP, ROC AUC, and Prec@$k$) show the distribution of performances across all 297 models via boxplots for each dataset. 
For most datasets, there exists \textbf{considerable difference between the best and the worst performing model}---suggesting that effective model selection would be beneficial.



\subsection{Corresponding Results based on Other Metrics}
\label{ssec:other}

Due to space limit, all performance results in Experiments (Sec. \ref{sec:experiment}) are based on Average Precision (\textbf{AP}). 
For completeness, we provide the results of the same analysis 
corresponding to \textbf{ROC} AUC and \textbf{Prec@}$\bm{k}$ metrics.

The conclusions are similar for these two metrics.

{\bf Cluster quality based methods.~} Specifically, Tables \ref{table:pval12roc} and \ref{table:pval12prn} present, resp. for \roc and \pre,  the pairwise comparison of cluster quality based methods and the baselines (\rand and \ifor).
Three strategies \rs, \ch, and \xb appear to stand out from others. However,
none of the methods are not significantly different from (and few are sometimes worse than) \rand.
Most of them are significantly worse than \ifor, with otherwise a very large $p$-value.

{\bf Other stand-alone methods.~} Tables \ref{table:emmv_roc} and \ref{table:emmv_prn} present, resp. for \roc and \pre, the pairwise comparison of all the stand-alone methods (only \rs, \ch, and \xb from above) and the baselines. We find that they are not different from each other or \rand---implying that \textbf{stand-alone model selection techniques would not be useful in practice}.

{\bf Consensus-based methods.~} Tables \ref{table:consensus_roc} and \ref{table:consensus_prn} show, resp. for \roc and \pre, that all  consensus-based techniques, namely \udr, \mc, \mcs, \hits, and \ens, are comparable to each other in terms of selection performance.

Finally, Tables \ref{table:qroc} and \ref{table:qprn} provide, resp. for \roc and \pre, a summary of the results for all the unsupervised model selection methods we studied. Main take-aways are: 
(1) \textbf{Consensus-based model selection methods are more competitive} than stand-alone methods, where all of them achieve \textbf{significantly better performance than \rand selection}.
(2) Further, they are most often not different from \ifor (a state-of-the-art detector) and sometimes even better (w.r.t. \roc). However, their absolute  difference (i.e. effect size) is negligible as shown in Figures \ref{fig:diffroc} and \ref{fig:diffprn}, resp. for \roc and \pre. Notably, their performance differences are not far from zero, suggesting that \textbf{consensus-based selection would also not be preferrable in practice}, since training a \textit{single} \ifor model is much faster over training a \textit{pool} of models (with considerable running time overhead) to select from.

\newpage

\begin{table}[!h]
	\footnotesize
\centering
	\caption{Comparison of cluster quality based methods and baselines
		 by one-sided paired Wilcoxon signed rank test on ROC AUC.
		$p$-values \textbf{bolded} (\underline{underlined}) highlight the cases where row-method  is significantly \textbf{better} (\underline{worse}) than col-method at $p$$\leq$$0.05$.
	} 
	\vspace{0.05in}
	   \scalebox{0.7}{
	   		\begin{tabular}{l|lllllll|ll}
	   			\toprule
	   		&	\std & \h &  \s &\iind & \db &  \sd & \dunn & \rnd & \ifr \\
	   		\midrule
	   		\xb,\rs,\ch & \textBF{0.001} & 0.407  &  \textBF{0.007} & 0.389 & 0.272    & 0.099 & 0.518 & 0.358   & \underline{0.980} \\
	   		\std &      & \underline{1.000} & \underline{0.990} & \underline{1.000} & \underline{0.994} &  \underline{0.995} & \underline{1.000} & \underline{1.000}     & \underline{1.000}     \\
	   		\h &      &      &  \textBF{0.021} & 0.500   & 0.487 &  0.320 & 0.831 & 0.818 & \underline{1.000} \\
	   		\s &      &            &     & \underline{0.974} & 0.816 & 0.704 & \underline{0.994} & \underline{0.994} & \underline{1.000} \\
	   		\iind &      &           &      &      & 0.487  & 0.323 & 0.849 & 0.821 & \underline{1.000} \\
	   		\db &      &           &      &      &        & 0.368 & 0.815 & 0.662 & \underline{0.996} \\
	   		\sd &   &           &      &      &      &      & 0.842 & 0.905  & \underline{0.999} \\
	   		\dunn &      &           &      &      &      &     &      & 0.110 & \underline{0.998}  \\
	   		\rnd &      &            &      &      &      &      &      &      & \underline{1.000}    
%
	   		\end{tabular}
}
	\label{table:pval12roc} 
\end{table}

\begin{table}[!h]
	\footnotesize
	\centering
	\caption{Comparison of stand-alone methods and baselines w.r.t. ROC AUC.} 
	\scalebox{0.9}{
		\begin{tabular}{l|lll|ll}
			\toprule
			&  \emmes	& \mv & \ireos & \rnd & \ifr \\
			\midrule
			\xb,\rs,\ch & 0.364 & 0.422 & 0.934 & 0.358   & \underline{0.980}  \\	
			\emmes & & 0.079 & \underline{0.969} & 0.358 &  \underline{0.992} \\
			\mv& & & \underline{0.977} & 0.369 &  \underline{0.997} \\
			\ireos & & & & \textBF{0.006} & 0.702  \\
			\bottomrule
		\end{tabular}
	}
	\label{table:emmv_roc} 
\end{table}

\begin{table}[!h]
	\footnotesize
	\centering
	\caption{Comparison of consensus-based methods (\udr, \mc, \mcs are based on ${NDCG}$) w.r.t. ROC AUC.} 
	\scalebox{0.9}{
		\begin{tabular}{l|llll}
			\toprule
			&   \mc  & \mcs & \hits & \ens \\ 
			\midrule
			\udr & 0.462 & 0.070 & 0.408 & 0.232 \\ 
			\mc  & & 0.100 &  0.134 &  0.069 \\ 
			\mcs  & & & 0.681 & 0.511 \\ 
			\hits  & & & & 0.740 \\ 
			\bottomrule
		\end{tabular}
	}
	\label{table:consensus_roc} 
\end{table}
 
\begin{table}[!h]
\centering
	\caption{\textbf{Summary of results}: $p$-values by one-sided paired Wilcoxon signed rank test comparing UOMS methods to the baselines, smallest $q$-th best model with no significant difference, and mean/standard deviation ROC AUC across datasets. 
	} 
	   \scalebox{0.72}{
	\begin{tabular}{cl|cc|r|cc} 
	    \toprule
		& \textbf{Method} &  \rand & \ifor & \qroc & mean ROC & std ROC \\ 
		\midrule
		\multirow{3}{*}{\rotatebox[origin=c]{90}{\parbox[c]{1.1cm}{\centering S-alone}}}
		& \xb,\rs,\ch    & 0.358 & \underline{0.980} &	138 & 0.690 & 0.206 \\ 
 		\cline{2-7}
 		&\emmes  &  0.358 &  \underline{0.992} & 142 & 0.682 & 0.216 \\ 
 		\cline{2-7}
 		&\ireos	& \textBF{0.006} & 0.702 & 83&  0.730 &  0.203 \\ 
 		 
 		\midrule
 		\multirow{11}{*}{\rotatebox[origin=c]{90}{\parbox[c]{4cm}{\centering Consensus-based}}}
	&	\udr-${\rho}$ & \textBF{0.000} & 0.279 & {82} & {0.763}& {0.180}  \\
	&	\udr-$\tau$ & \textBF{0.000} & 0.186 & 75	& 0.769	& 0.180	\\
	&	\udr-$NDCG$ & \textBF{0.000} & 0.175 &  75 & 0.769 & 0.183	\\
	\cline{2-7}
	&	\mc-${\rho}$& \textBF{0.000} & \textBF{0.036} & 92	& 0.767	& 0.168	\\
	&	\mc-${\tau}$& \textBF{0.000} & \textBF{0.011}	& {91}	& {0.769}	& {0.167} \\
	&	\mc-${NDCG}$& \textBF{0.000} & \textBF{0.034}	& 86 & 0.771 & 0.170
	\\
		\cline{2-7}
	&	\mcs-${\rho}$& \textBF{0.000} & 0.483 	& 100 & 0.763 & 0.173 \\	
	&	\mcs-${\tau}$& \textBF{0.000} & 0.121 & 94 & 0.761 & 0.167 \\
	&	\mcs-${NDCG}$&  \textBF{0.000} & {0.274}& 94 & 0.766 & 0.165 \\
 				
 			\cline{2-7}
 		&\hits& \textBF{0.000} & 0.148 & 97 & 0.762 & 0.169 \\
 		&\ens& \textBF{0.000} & 0.230 & 86 &0.749& 0.183 \\
 		 	\midrule
 		 	\multirow{2}{*}{\rotatebox[origin=c]{90}{\parbox[c]{0.6cm}{\centering Agg.}}}	
 		 	&\hitsc& \textBF{0.000} & \textBF{0.018} & 77 & 0.785& 0.163\\ 
 		 &\ensc& \textBF{0.000} & 0.135 & 87 & 0.749 & 0.184	 \\
 				\midrule	
 			\multirow{2}{*}{\rotatebox[origin=c]{90}{\parbox[c]{1cm}{\centering Base.}}}	
 		&\rand &\cdash 	& \underline{1.000} & 183 & 0.704 & 0.133\\ 
 		&\ifor 	&\cdash & \cdash & 102 & 0.763 & 0.166 \\ 
 		\bottomrule
	\end{tabular}}
	\label{table:qroc} 
\end{table}

\clearpage

\begin{table}[!h]
	\footnotesize
\centering
	\caption{Comparison of cluster quality based methods and baselines
		 by one-sided paired Wilcoxon signed rank test on {Prec@}${k}$.
		$p$-values \textbf{bolded} (\underline{underlined}) highlight the cases where row-method  is significantly \textbf{better} (\underline{worse}) than col-method at $p$$\leq$$0.05$.
	} 
	\vspace{0.05in}
	   \scalebox{0.7}{
	   		\begin{tabular}{l|lllllll|ll}
	   			\toprule
	   		&	\std & \h &  \s &\iind & \db &  \sd & \dunn & \rnd & \ifr \\
	   		\midrule
	   		\xb,\rs,\ch & \textBF{0.000} & 0.125  &  \textBF{0.031} & 0.109 & 0.173    & \textBF{0.026} & 0.274 & 0.090   & 0.716 \\
	   		\std &      & \underline{0.998} & \underline{0.985} & \underline{0.999} & \underline{0.989} &  \underline{0.956} & \underline{1.000} & \underline{1.000}     & \underline{1.000}     \\
	   		\h &      &      & 0.447 & 0.704   & 0.623 &  0.191 & 0.581 & 0.500 & \underline{0.967} \\
	   		\s &      &            &     & 0.488 & 0.815 & 0.313 & 0.875 & 0.757 & \underline{0.961} \\
	   		\iind &      &           &      &      & 0.631  & 0.203 & 0.632 & 0.544 & \underline{0.978} \\
	   		\db &      &           &      &      &        & 0.166 & 0.719 & 0.423 & 0.915 \\
	   		\sd &   &           &      &      &      &      & 0.929 & 0.879  & \underline{0.993} \\
	   		\dunn &      &           &      &      &      &     &      & 0.201 & 0.923  \\
	   		\rnd &      &            &      &      &      &      &      &      & \underline{0.999}    
%
	   		\end{tabular}
}
	\label{table:pval12prn} 
\end{table}

\begin{table}[!h]
	\footnotesize
	\centering
	\caption{Comparison of stand-alone methods and baselines w.r.t. {Prec@}${k}$.} 
	\scalebox{0.9}{
		\begin{tabular}{l|lll|ll}
			\toprule
			&  \emmes	& \mv & \ireos & \rnd & \ifr \\
			\midrule
			\xb,\rs,\ch & 0.272 & 0.193 & 0.405 & 0.090   & 0.716  \\	
			\emmes & & 0.187 & 0.696 & 0.730 &  \underline{0.967} \\
			\mv& & & 0.770 & 0.829 &  \underline{0.987} \\
			\ireos & & & & 0.423 & 0.944  \\
			\bottomrule
		\end{tabular}
	}
	\label{table:emmv_prn} 
\end{table}

\begin{table}[!h]
	\footnotesize
	\centering
	\caption{Comparison of consensus-based methods (\udr, \mc, \mcs are based on ${NDCG}$) w.r.t. {Prec@}${k}$.} 
	\scalebox{0.9}{
		\begin{tabular}{l|llll}
			\toprule
			&   \mc  & \mcs & \hits & \ens \\ 
			\midrule
			\udr & 0.645 & 0.403 & 0.464 & 0.296 \\ 
			\mc  & & 0.145 &  0.227 &  0.341\\ 
			\mcs  & & & 0.375 & 0.488 \\ 
			\hits  & & & & 0.608 \\ 
			\bottomrule
		\end{tabular}
	}
	\label{table:consensus_prn} 
\end{table}
 
\begin{table}[!h]
\centering
	\caption{\textbf{Summary of results}: $p$-values by one-sided paired Wilcoxon signed rank test comparing UOMS methods to the baselines, smallest $q$-th best model with no significant difference, and mean/standard deviation {Prec@}${k}$ across datasets. 
	} 
	   \scalebox{0.68}{
	\begin{tabular}{cl|cc|r|cc} 
	    \toprule
		& \textbf{Method} &  \rand & \ifor & \qprn & mean {Prec@}${k}$ & std {Prec@}${k}$ \\ 
		\midrule
		\multirow{3}{*}{\rotatebox[origin=c]{90}{\parbox[c]{1.1cm}{\centering S-alone}}}
		& \xb,\rs,\ch    & 0.090 & 0.716 &	91 & 0.348 & 0.277 \\ 
 		\cline{2-7}
 		&\emmes  &  0.730 &  \underline{0.967} & 119 & 0.303 & 0.254 \\ 
 		\cline{2-7}
 		&\ireos	 & 0.423 & 0.944 & 102 & 0.316 & 0.255\\ 
 		
 		 
 		\midrule
 		\multirow{11}{*}{\rotatebox[origin=c]{90}{\parbox[c]{4cm}{\centering Consensus-based}}}
	&	\udr-${\rho}$ & \textBF{0.039} & \underline{0.965} & {115} & {0.354} & {0.271}  \\
	&	\udr-$\tau$ & \textBF{0.025} & 0.942 & 110	& 0.356 & 0.263	\\
	&	\udr-$NDCG$ & \textBF{0.002} & {0.600} &  86 & 0.372 & 0.255	\\
	\cline{2-7}
	&	\mc-${\rho}$& \textBF{0.002} & 0.555 & 98 & 0.369	& 0.271	\\
	&	\mc-${\tau}$& \textBF{0.002} & 0.833	& {103}	& {0.370}	& {0.280} \\
	&	\mc-${NDCG}$& \textBF{0.000} & {0.228}	& 89 & 0.378 & 0.270
	\\
		\cline{2-7}
	&	\mcs-${\rho}$& \textBF{0.008} & 0.937 	& 115 & 0.361 & 0.276 \\	
	&	\mcs-${\tau}$& \textBF{0.002} & 0.595 & 96 & 0.374 & 0.290 \\
	&	\mcs-${NDCG}$&  \textBF{0.002} & {0.210}& 92 & 0.367 & 0.274 \\
 				
 			\cline{2-7}
 		&\hits& \textBF{0.001} & 0.583 & 99 & 0.376 & 0.280 \\
 		&\ens& \textBF{0.004} & 0.595 & 92 &0.351& 0.261 \\
 		 	\midrule
 		 	\multirow{2}{*}{\rotatebox[origin=c]{90}{\parbox[c]{0.6cm}{\centering Agg.}}}	
 		 	&\hitsc& \textBF{0.000} & 0.293 & 89 & 0.380 & 0.263\\ 
 		 &\ensc& \textBF{0.005} & 0.722 & 89 & 0.350 & 0.262	 \\
 				\midrule	
 			\multirow{2}{*}{\rotatebox[origin=c]{90}{\parbox[c]{1cm}{\centering Base.}}}	
 		&\rand &\cdash 	& \underline{0.999} & 153 & 0.325 & 0.217\\ 
 		&\ifor 	&\cdash & \cdash & 91 & 0.374 & 0.280 \\ 
 		\bottomrule
	\end{tabular}}
	\label{table:qprn} 
\end{table}


\begin{figure}[!t]
	\begin{center}
		\centerline{\includegraphics[width=0.95\columnwidth]{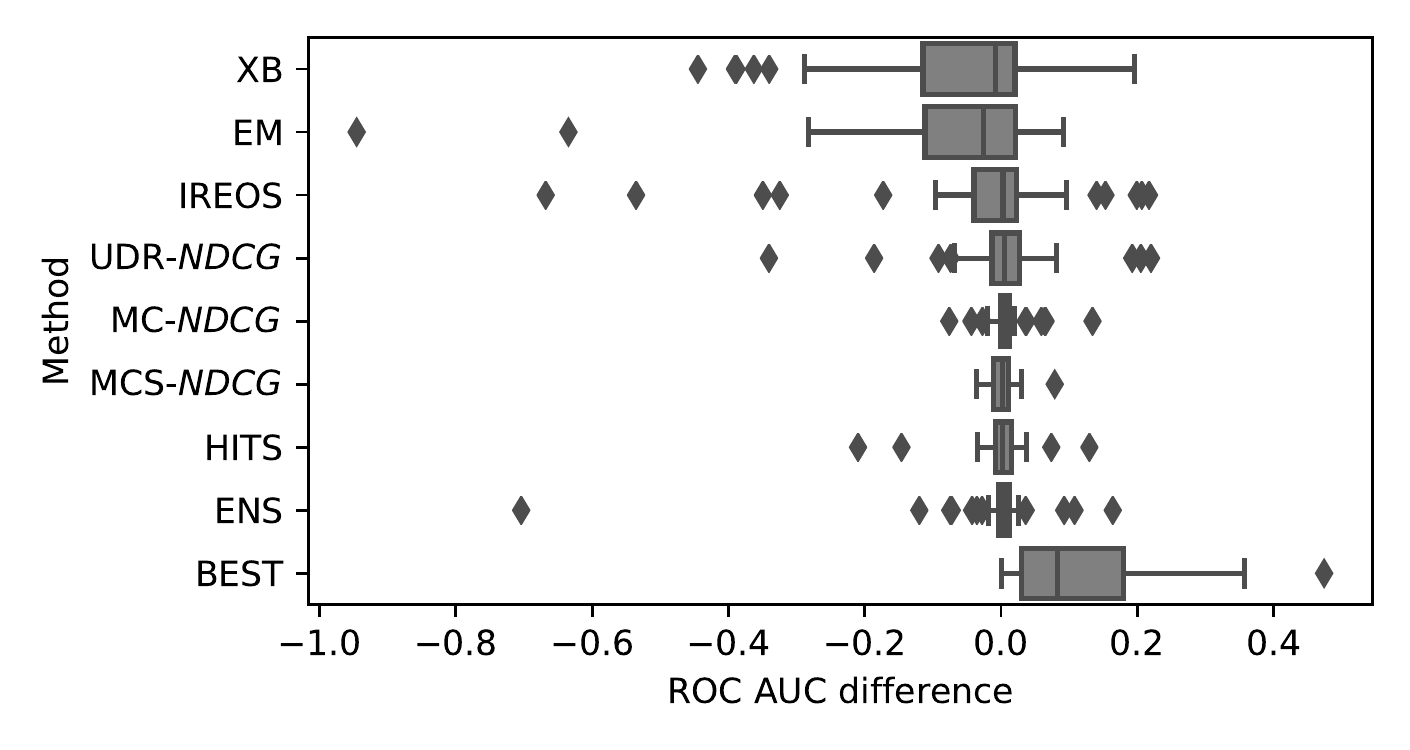}}
		\caption{Distribution across datasets of performance difference: ROC AUC of selected model (by each UOMS method studied) minus that of \ifor. Stand-alone methods and \udr are subpar, whereas other consensus-based differences concentrate around zero (not notably different from \ifor). Also shown for comparison is \best model on each dataset. 
		}
		\label{fig:diffroc}
	\end{center}
\end{figure}

\begin{figure}[!t]
	\begin{center}
	\centerline{\includegraphics[width=0.95\columnwidth]{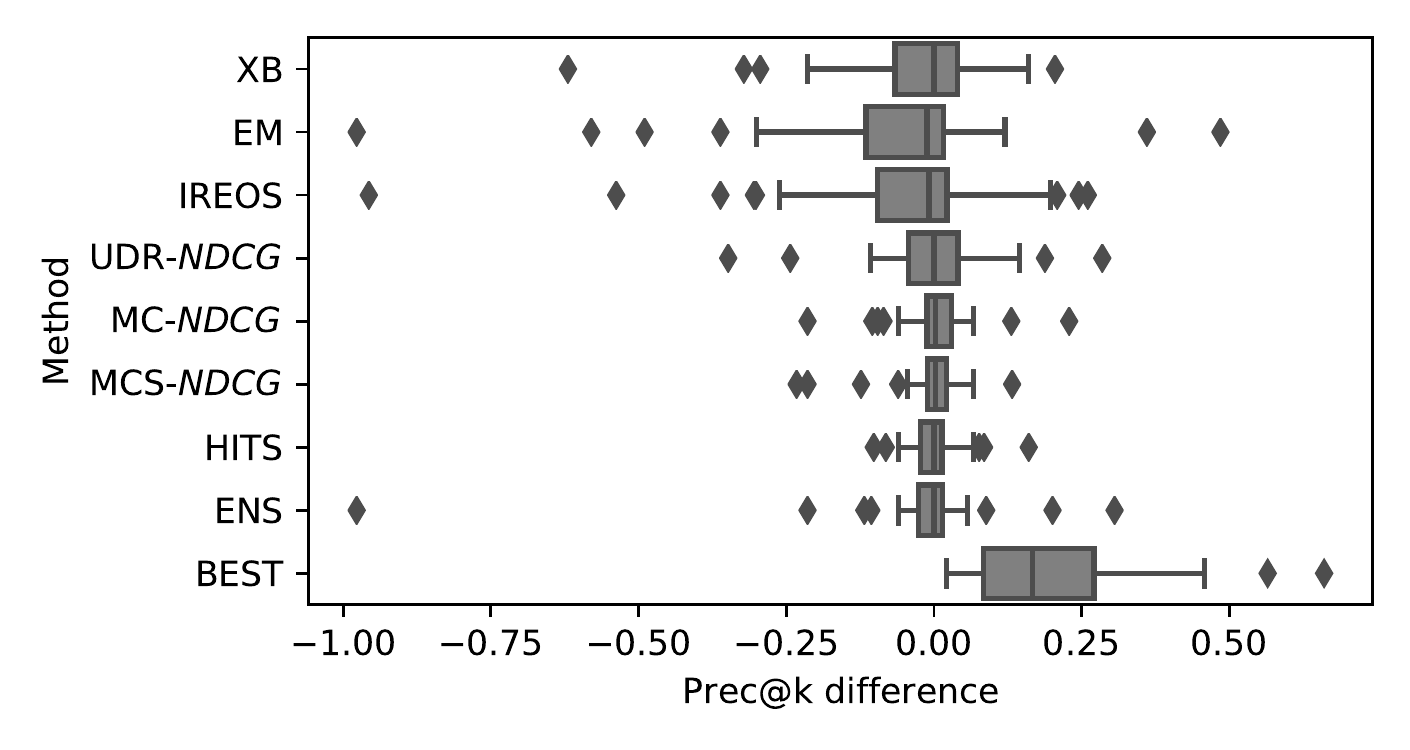}}
		\caption{Distribution across datasets of  performance difference: Prec@$k$ of selected model (by each UOMS method studied) minus that of \ifor. Stand-alone methods and \udr are subpar, whereas other consensus-based differences concentrate around zero (not notably different from \ifor). Also shown for comparison is \best model on each dataset. 
		}
		\label{fig:diffprn}
	\end{center}
\end{figure}

\clearpage

\begin{table*}[!h]
	\footnotesize
\centering
	\caption{Family-wise model performance in AP. Values in \textbf{bold} highlight the model that outperforms for each dataset (per row). iForest achieves the highest average performance across all datasets.   
	 } 
	\vspace{0.1in}	
	   \scalebox{0.9}{
		\begin{tabular}{l|llllllll}
			\toprule
			\textbf{Dataset}                 & \textbf{LODA} & \textbf{ABOD}  & \textbf{iForest} & \textbf{kNN}   & \textbf{LOF} & \textbf{HBOS}  & \textbf{OCSVM} & \textbf{COF}   \\
			\midrule
			\textbf{annthyroid (ODDS)}       & 0.136         & 0.232          & 0.340            & 0.228          & 0.172        & \textbf{0.388} & 0.145          & 0.138          \\
			\textbf{arrhythmia (ODDS)}       & 0.387         & 0.315          & \textbf{0.470}   & 0.392          & 0.362        & 0.431          & 0.250          & 0.404          \\
			\textbf{breastw (ODDS)}          & 0.964         & 0.702          & \textbf{0.972}   & 0.942          & 0.331        & 0.959          & 0.544          & 0.304          \\
			\textbf{glass (ODDS)}            & 0.063         & 0.137          & 0.104            & 0.106          & 0.117        & 0.061          & 0.063          & \textbf{0.154} \\
			\textbf{ionosphere (ODDS)}       & 0.766         & \textbf{0.921} & 0.784            & 0.868          & 0.819        & 0.288          & 0.492          & 0.852          \\
			\textbf{letter (ODDS)}           & 0.092         & 0.319          & 0.089            & 0.258          & 0.359        & 0.080          & 0.138          & \textbf{0.459} \\
			\textbf{lympho (ODDS)}           & 0.447         & 0.555          & \textbf{0.957}   & 0.763          & 0.668        & 0.905          & 0.418          & 0.464          \\
			\textbf{mammography (ODDS)}      & 0.218         & 0.147          & \textbf{0.234}   & 0.169          & 0.102        & 0.096          & 0.156          & 0.064          \\
			\textbf{mnist (ODDS)}            & 0.203         & 0.329          & 0.261            & \textbf{0.401} & 0.273        & 0.097          & 0.204          & 0.195          \\
			\textbf{musk (ODDS)}             & 0.904         & 0.038          & 0.990            & 0.588          & 0.130        & \textbf{0.997} & 0.498          & 0.174          \\
			\textbf{optdigits (ODDS)}        & 0.025         & 0.057          & 0.049            & 0.021          & 0.037        & \textbf{0.177} & 0.031          & 0.048          \\
			\textbf{pendigits (ODDS)}        & 0.245         & 0.057          & \textbf{0.280}   & 0.104          & 0.038        & 0.231          & 0.086          & 0.037          \\
			\textbf{pima (ODDS)}             & 0.445         & 0.508          & 0.492            & \textbf{0.524} & 0.441        & 0.521          & 0.385          & 0.429          \\
			\textbf{satellite (ODDS)}        & 0.630         & 0.430          & 0.664            & 0.562          & 0.375        & \textbf{0.711} & 0.456          & 0.368          \\
			\textbf{satimage-2 (ODDS)}       & 0.904         & 0.212          & \textbf{0.916}   & 0.615          & 0.055        & 0.717          & 0.486          & 0.078          \\
			\textbf{speech (ODDS)}           & 0.018         & \textbf{0.093} & 0.020            & 0.024          & 0.031        & 0.025          & 0.022          & 0.034          \\
			\textbf{thyroid (ODDS)}          & 0.238         & 0.218          & 0.587            & 0.354          & 0.157        & \textbf{0.630} & 0.196          & 0.032          \\
			\textbf{vertebral (ODDS)}        & 0.089         & 0.098          & 0.094            & 0.090          & 0.101        & 0.087          & \textbf{0.131} & 0.116          \\
			\textbf{vowels (ODDS)}           & 0.140         & \textbf{0.690} & 0.134            & 0.487          & 0.348        & 0.083          & 0.080          & 0.408          \\
			\textbf{wbc (ODDS)}              & 0.603         & 0.367          & 0.599            & 0.533          & 0.497        & \textbf{0.673} & 0.321          & 0.261          \\
			\textbf{wine (ODDS)}             & 0.286         & 0.082          & 0.215            & 0.253          & 0.253        & \textbf{0.402} & 0.249          & 0.081          \\
			\textbf{Annthyroid (DAMI)}       & 0.097         & 0.137          & \textbf{0.160}   & 0.126          & 0.134        & 0.145          & 0.079          & 0.130          \\
			\textbf{Arrhythmia (DAMI)}       & 0.685         & 0.668          & \textbf{0.757}   & 0.711          & 0.702        & 0.745          & 0.523          & 0.712          \\
			\textbf{Cardiotocography (DAMI)} & 0.433         & 0.254          & \textbf{0.433}   & 0.316          & 0.280        & 0.344          & 0.314          & 0.267          \\
			\textbf{HeartDisease (DAMI)}     & 0.562         & 0.547          & 0.538            & 0.557          & 0.509        & \textbf{0.619} & 0.475          & 0.486          \\
			\textbf{InternetAds (DAMI)}      & 0.251         & 0.293          & 0.490            & 0.289          & 0.263        & \textbf{0.521} & 0.237          & 0.261          \\
			\textbf{PageBlocks (DAMI)}       & 0.464         & 0.416          & 0.449            & \textbf{0.526} & 0.360        & 0.201          & 0.268          & 0.232          \\
			\textbf{Pima (DAMI)}             & 0.448         & 0.506          & 0.494            & \textbf{0.529} & 0.467        & 0.487          & 0.392          & 0.432          \\
			\textbf{SpamBase (DAMI)}         & 0.370         & 0.357          & 0.487            & 0.406          & 0.364        & \textbf{0.532} & 0.366          & 0.392          \\
			\textbf{Stamps (DAMI)}           & 0.332         & 0.218          & \textbf{0.336}   & 0.313          & 0.228        & 0.315          & 0.209          & 0.159          \\
			\textbf{Wilt (DAMI)}             & 0.039         & 0.065          & 0.045            & 0.053          & 0.075        & 0.044          & 0.065          & \textbf{0.101} \\
			\textbf{ALOI (DAMI)}             & 0.034         & 0.102          & 0.033            & 0.057          & 0.100        & 0.031          & 0.035          & \textbf{0.144} \\
			\textbf{Glass (DAMI)}            & 0.085         & \textbf{0.221} & 0.183            & 0.146          & 0.118        & 0.115          & 0.107          & 0.179          \\
			\textbf{PenDigits (DAMI)}        & 0.003         & 0.031          & 0.005            & \textbf{0.040} & 0.014        & 0.004          & 0.016          & 0.017          \\
			\textbf{Shuttle (DAMI)}          & 0.111         & 0.250          & 0.071            & \textbf{0.326} & 0.296        & 0.094          & 0.095          & 0.173          \\
			\textbf{Waveform (DAMI)}         & 0.052         & 0.055          & 0.057            & \textbf{0.115} & 0.095        & 0.053          & 0.069          & 0.102          \\
			\textbf{WBC (DAMI)}              & 0.743         & 0.595          & \textbf{0.858}   & 0.671          & 0.359        & 0.683          & 0.424          & 0.146          \\
			\textbf{WDBC (DAMI)}             & 0.720         & 0.296          & 0.669            & 0.571          & 0.554        & \textbf{0.725} & 0.322          & 0.295          \\
			\textbf{WPBC (DAMI)}             & 0.235         & 0.231          & 0.229            & 0.233          & 0.230        & \textbf{0.239} & 0.237          & 0.219          \\
			\midrule
			\textbf{average}                 & 0.345          & 0.301          & \textbf{0.399}   & 0.366         & 0.277        & 0.371          & 0.246          & 0.245          \\
			\textbf{STD}                     & 0.282          & 0.220          & 0.304            & 0.248         & 0.199        & 0.295          & 0.165          & 0.188          \\
			\bottomrule
		\end{tabular}
}
	\label{table:model_performance_ap} 
\end{table*}


\begin{table*}[!h]
	\footnotesize
\centering
	\caption{Family-wise model performance in ROC AUC. Values in \textbf{bold} highlight the model that outperforms for each dataset (per row). kNN (0.764) and iForest (0.763) achieve the highest average performance across all datasets.   
	 } 
	\vspace{0.1in}	
	   \scalebox{0.9}{
		\begin{tabular}{l|llllllll}
			\toprule
			\textbf{Dataset}                 & \textbf{LODA} & \textbf{ABOD}  & \textbf{iForest} & \textbf{kNN}   & \textbf{LOF} & \textbf{HBOS}  & \textbf{OCSVM} & \textbf{COF}   \\
			\midrule
			\textbf{annthyroid (ODDS)}       & 0.572         & 0.823          & \textbf{0.841}   & 0.775          & 0.729        & \textbf{0.736} & 0.517          & 0.689          \\
			\textbf{arrhythmia (ODDS)}       & 0.735         & 0.751          & \textbf{0.803}   & 0.777          & 0.764        & \textbf{0.806} & 0.522          & 0.757          \\
			\textbf{breastw (ODDS)}          & 0.980         & 0.898          & \textbf{0.988}   & 0.980          & 0.500        & 0.985          & 0.481          & 0.459          \\
			\textbf{glass (ODDS)}            & 0.539         & 0.766          & 0.707            & 0.747          & 0.747        & 0.638          & 0.429          & \textbf{0.772} \\
			\textbf{ionosphere (ODDS)}       & 0.814         & \textbf{0.928} & 0.838            & 0.898          & 0.870        & 0.357          & 0.548          & 0.879          \\
			\textbf{letter (ODDS)}           & 0.584         & \textbf{0.880} & 0.629            & 0.842          & 0.846        & 0.581          & 0.554          & \textbf{0.880} \\
			\textbf{lympho (ODDS)}           & 0.814         & 0.936          & \textbf{0.998}   & 0.971          & 0.938        & 0.985          & 0.607          & 0.834          \\
			\textbf{mammography (ODDS)}      & 0.854         & 0.822          & \textbf{0.862}   & 0.845          & 0.729        & 0.799          & 0.629          & 0.700          \\
			\textbf{mnist (ODDS)}            & 0.586         & 0.797          & 0.794            & \textbf{0.856} & 0.708        & 0.515          & 0.536          & 0.615          \\
			\textbf{musk (ODDS)}             & 0.991         & 0.072          & 0.999            & 0.830          & 0.521        & \textbf{1.000} & 0.669          & 0.534          \\
			\textbf{optdigits (ODDS)}        & 0.414         & 0.477          & 0.713            & 0.383          & 0.463        & \textbf{0.877} & 0.463          & 0.526          \\
			\textbf{pendigits (ODDS)}        & 0.934         & 0.692          & \textbf{0.948}   & 0.818          & 0.516        & 0.921          & 0.548          & 0.508          \\
			\textbf{pima (ODDS)}             & 0.629         & 0.685          & 0.652            & \textbf{0.717} & 0.630        & 0.634          & 0.497          & 0.583          \\
			\textbf{satellite (ODDS)}        & 0.644         & 0.594          & 0.703            & 0.703          & 0.546        & \textbf{0.785} & 0.506          & 0.519          \\
			\textbf{satimage-2 (ODDS)}       & 0.988         & 0.854          & \textbf{0.993}   & 0.965          & 0.678        & 0.973          & 0.610          & 0.537          \\
			\textbf{speech (ODDS)}           & 0.474         & \textbf{0.688} & 0.473            & 0.500          & 0.525        & 0.473          & 0.492          & 0.584          \\
			\textbf{thyroid (ODDS)}          & 0.820         & 0.945          & \textbf{0.983}   & 0.960          & 0.771        & \textbf{0.950} & 0.550          & 0.581          \\
			\textbf{vertebral (ODDS)}        & 0.315         & 0.375          & 0.349            & 0.333          & 0.380        & 0.297          & \textbf{0.482} & 0.454          \\
			\textbf{vowels (ODDS)}           & 0.712         & \textbf{0.976} & 0.736            & 0.944          & 0.905        & 0.676          & 0.529          & 0.877          \\
			\textbf{wbc (ODDS)}              & 0.941         & 0.918          & 0.938            & 0.935          & 0.892        & \textbf{0.950} & 0.603          & 0.792          \\
			\textbf{wine (ODDS)}             & 0.853         & 0.490          & 0.794            & 0.779          & 0.758        & \textbf{0.873} & 0.536          & 0.373          \\
			\textbf{Annthyroid (DAMI)}       & 0.491         & \textbf{0.717} & \textbf{0.679}   & 0.658          & 0.679        & 0.646          & 0.471          & 0.666          \\
			\textbf{Arrhythmia (DAMI)}       & 0.687         & 0.725          & \textbf{0.750}   & 0.736          & 0.732        & 0.736          & 0.506          & 0.736          \\
			\textbf{Cardiotocography (DAMI)} & 0.689         & 0.458          & \textbf{0.689}   & 0.503          & 0.544        & 0.566          & 0.489          & 0.522          \\
			\textbf{HeartDisease (DAMI)}     & 0.608         & 0.612          & 0.602            & 0.637          & 0.582        & \textbf{0.670} & 0.502          & 0.542          \\
			\textbf{InternetAds (DAMI)}      & 0.548         & 0.657          & 0.690            & 0.626          & 0.587        & \textbf{0.695} & 0.499          & 0.579          \\
			\textbf{PageBlocks (DAMI)}       & 0.785         & 0.780          & \textbf{0.894}   & \textbf{0.889} & 0.759        & 0.679          & 0.558          & 0.610          \\
			\textbf{Pima (DAMI)}             & 0.624         & 0.666          & 0.644            & \textbf{0.706} & 0.650        & 0.594          & 0.504          & 0.587          \\
			\textbf{SpamBase (DAMI)}         & 0.433         & 0.403          & 0.635            & 0.535          & 0.441        & \textbf{0.676} & 0.463          & 0.450          \\
			\textbf{Stamps (DAMI)}           & 0.891         & 0.793          & \textbf{0.901}   & 0.872          & 0.702        & 0.876          & 0.582          & 0.541          \\
			\textbf{Wilt (DAMI)}             & 0.363         & 0.628          & 0.457            & 0.538          & 0.626        & 0.419          & 0.489          & \textbf{0.695} \\
			\textbf{ALOI (DAMI)}             & 0.504         & 0.739          & 0.534            & 0.641          & 0.744        & 0.508          & 0.506          & \textbf{0.796} \\
			\textbf{Glass (DAMI)}            & 0.659         & \textbf{0.854} & 0.794            & 0.822          & 0.748        & 0.795          & 0.485          & 0.774          \\
			\textbf{PenDigits (DAMI)}        & 0.628         & 0.936          & 0.768            & \textbf{0.967} & 0.821        & 0.734          & 0.537          & 0.718          \\
			\textbf{Shuttle (DAMI)}          & 0.637         & 0.927          & 0.853            & \textbf{0.963} & 0.911        & 0.842          & 0.566          & 0.848          \\
			\textbf{Waveform (DAMI)}         & 0.664         & 0.666          & 0.707            & \textbf{0.743} & 0.716        & 0.703          & 0.492          & 0.689          \\
			\textbf{WBC (DAMI)}              & 0.983         & 0.954          & \textbf{0.991}   & 0.979          & 0.842        & 0.985          & 0.611          & 0.703          \\
			\textbf{WDBC (DAMI)}             & 0.945         & 0.890          & 0.936            & 0.924          & 0.871        & \textbf{0.963} & 0.629          & 0.800          \\
			\textbf{WPBC (DAMI)}             & 0.509         & 0.501          & 0.498            & 0.509          & 0.503        & \textbf{0.536} & 0.485          & 0.463          \\
			\midrule
			\textbf{average}                 & 0.688         & 0.725          & 0.763   & \textbf{0.764} & 0.689        & 0.729          & 0.530          & 0.645          \\
			\textbf{STD}                     & 0.188         & 0.197          & 0.168            & 0.175          & 0.146        & 0.188          & 0.054          & 0.138          \\
			\bottomrule
		\end{tabular}
}
	\label{table:model_performance_roc} 
\end{table*}

\clearpage

\begin{table*}[!h]
	\footnotesize
\centering
	\caption{Family-wise model performance in {Prec@}${k}$. Values in \textbf{bold} highlight the model that outperforms for each dataset (per row). iForest achieves the highest average performance across all datasets.   
	 } 
	\vspace{0.1in}	
	   \scalebox{0.9}{
		\begin{tabular}{l|llllllll}
			\toprule
			\textbf{Dataset}                 & \textbf{LODA} & \textbf{ABOD}  & \textbf{iForest} & \textbf{kNN}   & \textbf{LOF} & \textbf{HBOS}  & \textbf{OCSVM} & \textbf{COF}   \\
			\midrule
			\textbf{annthyroid (ODDS)}       & 0.180          & 0.301          & 0.337            & 0.297          & 0.209        & \textbf{0.387} & 0.180          & 0.169          \\
			\textbf{arrhythmia (ODDS)}       & 0.403          & 0.372          & 0.481            & 0.411          & 0.386        & \textbf{0.495} & 0.237          & 0.407          \\
			\textbf{breastw (ODDS)}          & 0.924          & 0.788          & 0.929            & 0.923          & 0.271        & \textbf{0.938} & 0.445          & 0.152          \\
			\textbf{glass (ODDS)}            & 0.019          & 0.111          & 0.111            & 0.111          & 0.136        & 0.014          & 0.040          & \textbf{0.143} \\
			\textbf{ionosphere (ODDS)}       & 0.645          & \textbf{0.849} & 0.648            & 0.753          & 0.725        & 0.228          & 0.439          & 0.764          \\
			\textbf{letter (ODDS)}           & 0.100          & 0.354          & 0.092            & 0.312          & 0.358        & 0.080          & 0.140          & \textbf{0.440} \\
			\textbf{lympho (ODDS)}           & 0.401          & 0.476          & \textbf{0.881}   & 0.639          & 0.560        & 0.808          & 0.347          & 0.405          \\
			\textbf{mammography (ODDS)}      & \textbf{0.286} & 0.197          & 0.261            & 0.251          & 0.194        & 0.114          & 0.192          & 0.114          \\
			\textbf{mnist (ODDS)}            & 0.212          & 0.376          & 0.293            & \textbf{0.420} & 0.315        & 0.095          & 0.218          & 0.246          \\
			\textbf{musk (ODDS)}             & 0.873          & 0.035          & 0.977            & 0.546          & 0.134        & \textbf{0.981} & 0.491          & 0.218          \\
			\textbf{optdigits (ODDS)}        & 0.001          & 0.045          & 0.025            & 0.000          & 0.029        & \textbf{0.211} & 0.018          & 0.067          \\
			\textbf{pendigits (ODDS)}        & 0.324          & 0.077          & \textbf{0.365}   & 0.110          & 0.072        & 0.269          & 0.113          & 0.063          \\
			\textbf{pima (ODDS)}             & 0.466          & 0.530          & 0.504            & \textbf{0.551} & 0.463        & 0.476          & 0.361          & 0.423          \\
			\textbf{satellite (ODDS)}        & 0.533          & 0.417          & 0.573            & 0.511          & 0.379        & \textbf{0.619} & 0.382          & 0.361          \\
			\textbf{satimage-2 (ODDS)}       & \textbf{0.865} & 0.260          & 0.862            & 0.577          & 0.086        & 0.661          & 0.465          & 0.145          \\
			\textbf{speech (ODDS)}           & 0.019          & \textbf{0.138} & 0.031            & 0.039          & 0.045        & 0.032          & 0.039          & 0.049          \\
			\textbf{thyroid (ODDS)}          & 0.287          & 0.198          & 0.620            & 0.332          & 0.149        & \textbf{0.645} & 0.224          & 0.000          \\
			\textbf{vertebral (ODDS)}        & 0.011          & 0.043          & 0.044            & 0.018          & 0.056        & 0.012          & 0.074          & \textbf{0.090} \\
			\textbf{vowels (ODDS)}           & 0.194          & \textbf{0.641} & 0.175            & 0.474          & 0.333        & 0.121          & 0.094          & 0.429          \\
			\textbf{wbc (ODDS)}              & 0.558          & 0.361          & 0.536            & 0.496          & 0.475        & \textbf{0.614} & 0.324          & 0.293          \\
			\textbf{wine (ODDS)}             & 0.257          & 0.000          & 0.140            & 0.194          & 0.203        & \textbf{0.408} & 0.200          & 0.043          \\
			\textbf{Annthyroid (DAMI)}       & 0.116          & 0.153          & \textbf{0.213}   & 0.134          & 0.165        & 0.191          & 0.074          & 0.162          \\
			\textbf{Arrhythmia (DAMI)}       & 0.604          & 0.630          & \textbf{0.655}   & 0.637          & 0.643        & 0.632          & 0.459          & 0.652          \\
			\textbf{Cardiotocography (DAMI)} & \textbf{0.407} & 0.266          & 0.396            & 0.311          & 0.288        & 0.303          & 0.259          & 0.264          \\
			\textbf{HeartDisease (DAMI)}     & 0.530          & 0.520          & 0.503            & 0.535          & 0.506        & \textbf{0.591} & 0.447          & 0.470          \\
			\textbf{InternetAds (DAMI)}      & 0.267          & 0.344          & 0.449            & 0.334          & 0.304        & \textbf{0.466} & 0.244          & 0.284          \\
			\textbf{PageBlocks (DAMI)}       & 0.458          & 0.425          & 0.397            & \textbf{0.506} & 0.376        & 0.158          & 0.264          & 0.268          \\
			\textbf{Pima (DAMI)}             & 0.476          & 0.512          & 0.499            & \textbf{0.547} & 0.485        & 0.448          & 0.369          & 0.421          \\
			\textbf{SpamBase (DAMI)}         & 0.351          & 0.359          & 0.518            & 0.421          & 0.338        & \textbf{0.562} & 0.357          & 0.382          \\
			\textbf{Stamps (DAMI)}           & 0.275          & 0.189          & 0.286            & 0.211          & 0.169        & \textbf{0.385} & 0.197          & 0.180          \\
			\textbf{Wilt (DAMI)}             & 0.001          & 0.012          & 0.012            & 0.003          & 0.058        & 0.006          & 0.043          & \textbf{0.121} \\
			\textbf{ALOI (DAMI)}             & 0.050          & 0.144          & 0.028            & 0.086          & 0.146        & 0.028          & 0.043          & \textbf{0.187} \\
			\textbf{Glass (DAMI)}            & 0.027          & 0.143          & 0.111            & 0.111          & 0.133        & 0.044          & 0.056          & \textbf{0.159} \\
			\textbf{PenDigits (DAMI)}        & 0.000          & 0.036          & 0.000            & 0.000          & 0.019        & 0.000          & 0.010          & \textbf{0.036} \\
			\textbf{Shuttle (DAMI)}          & 0.120          & \textbf{0.319} & 0.079            & 0.277          & 0.169        & 0.092          & 0.092          & 0.231          \\
			\textbf{Waveform (DAMI)}         & 0.057          & 0.069          & 0.065            & \textbf{0.191} & 0.161        & 0.063          & 0.083          & 0.143          \\
			\textbf{WBC (DAMI)}              & 0.630          & 0.429          & \textbf{0.723}   & 0.644          & 0.328        & 0.713          & 0.356          & 0.086          \\
			\textbf{WDBC (DAMI)}             & \textbf{0.650} & 0.271          & 0.633            & 0.592          & 0.536        & 0.648          & 0.350          & 0.286          \\
			\textbf{WPBC (DAMI)}             & 0.166          & 0.164          & 0.146            & 0.160          & 0.172        & \textbf{0.206} & 0.202          & 0.161          \\
			\midrule
			\textbf{average}                 & 0.327          & 0.296          & \textbf{0.374}   & 0.350          & 0.271        & 0.352          & 0.229          & 0.244          \\
			\textbf{STD}                     & 0.262          & 0.214          & 0.284            & 0.235          & 0.180        & 0.283          & 0.149          & 0.171          \\
			\bottomrule
		\end{tabular}
}
	\label{table:model_performance_prn} 
\end{table*}

\clearpage

\begin{figure*}[!t]
	\begin{center}
		\centerline{\includegraphics[width=2.1\columnwidth]{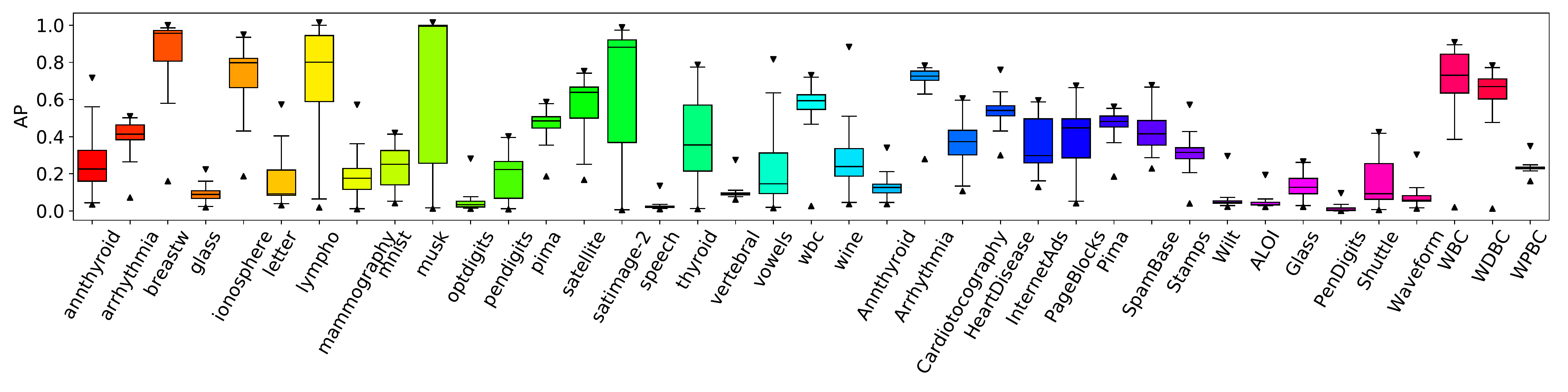}}
		\caption{Model performance boxplot (AP) for all datasets, where triangles mark the min and max. Model performance varies significantly for most datasets, showing the importance of model selection. 
		}
		\label{fig:APmodel}
	\end{center}
\end{figure*}

\begin{figure*}[!t]
	\begin{center}
		\centerline{\includegraphics[width=2.1\columnwidth]{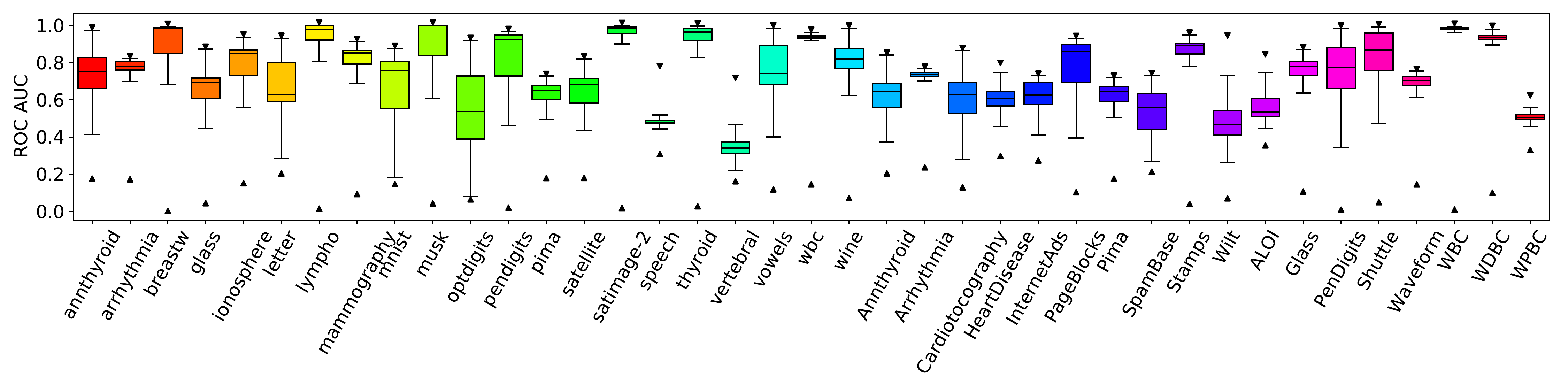}}
		\caption{Model performance boxplot (ROC AUC) for all datasets, where triangles mark the min and max. Model performance varies significantly for most datasets, showing the importance of model selection.}
		\label{fig:ROCmodel}
	\end{center}
\end{figure*}

\begin{figure*}[!t]
	\begin{center}
		\centerline{\includegraphics[width=2.1\columnwidth]{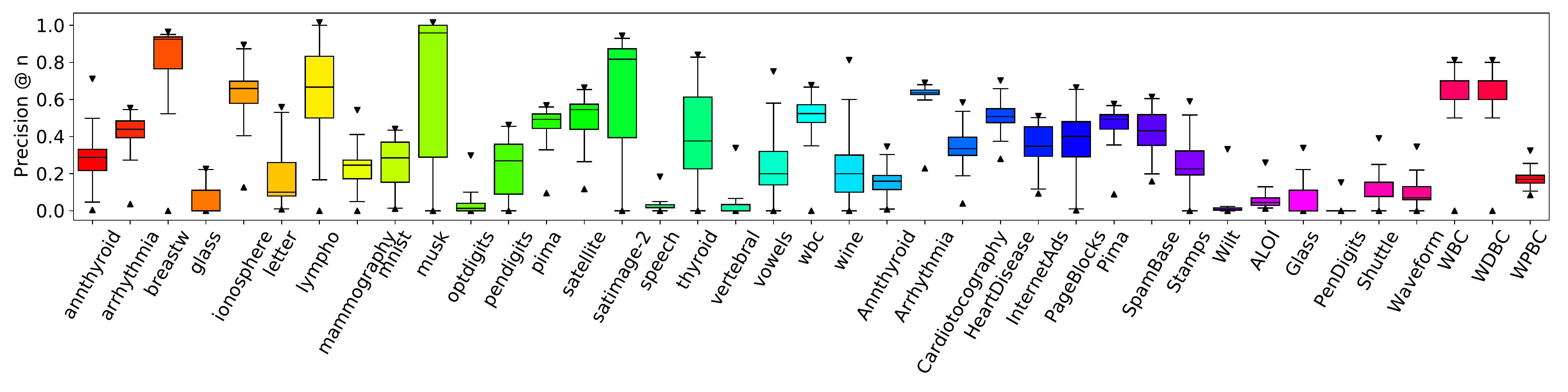}}
		\caption{Model performance boxplot ({Prec@}${k}$) for all datasets, where triangles mark the min and max. Model performance varies significantly for most datasets, showing the importance of model selection.}
		\label{fig:PRNmodel}
	\end{center}
\end{figure*}

\hide{
\begin{figure}[!t]
	\begin{center}
		\centerline{\includegraphics[width=0.8\columnwidth]{FIG/ROC-box-cluster}}
		\caption{.}
		\label{fig:ROCcluster}
	\end{center}
\end{figure}

\begin{figure}[!t]
	\begin{center}
		\centerline{\includegraphics[width=0.8\columnwidth]{FIG/PRN-box-cluster}}
		\caption{.}
		\label{fig:PRNcluster}
	\end{center}
\end{figure}

}


\end{document}
